\documentclass[journal]{IEEEtran}
\usepackage{cite}
\usepackage{amsmath,amssymb,amsfonts}
\usepackage{algorithmic}
\usepackage{graphicx}
\usepackage{textcomp}
\usepackage[dvipsnames]{xcolor}

\def\BibTeX{{\rm B\kern-.05em{\sc i\kern-.025em b}\kern-.08em
    T\kern-.1667em\lower.7ex\hbox{E}\kern-.125emX}}

\usepackage{hyperref}
\usepackage{graphicx}
\usepackage{amsmath}
\usepackage{algorithmic}
\usepackage{subcaption}
\usepackage{array, booktabs}
\usepackage{balance}
\usepackage{cite}
\newcolumntype{P}[1]{>{\centering\arraybackslash}p{#1}}
\newcolumntype{M}[1]{>{\centering\arraybackslash}m{#1}}
\usepackage{adjustbox}
\usepackage{multirow}

\usepackage{gensymb}
\usepackage{threeparttable}
\usepackage{caption}
\usepackage{subcaption}

\usepackage{caption}
\usepackage{float}

\begin{document}
%
\title{Dual-CyCon Net: A Cycle Consistent Dual-Domain Convolutional Neural Network Framework for Detection of Partial Discharge}
%
%
%

\author{Mohammad~Zunaed, 
        Ankur~Nath,
        and~Md.~Saifur~Rahman,
\thanks{Mohammad Zunaed, and Md. Saifur Rahman are with the Department of Electrical and Electronic Engineering, Bangladesh University of Engineering and Technology, Dhaka-1205, Bangladesh. Ankur Nath is with the Texas A{\&}M University, Department of Computer Science, College Station, Texas, United States of America. (e-mails: rafizunaed@gmail.com, anath@tamu.edu, saifur@eee.buet.ac.bd.)}
}

\maketitle

\begin{abstract}
In the last decade, researchers have been investigating the severity of insulation breakdown caused by partial discharge (PD) in overhead transmission lines with covered conductors or electrical equipment such as generators and motors used in various industries. Developing an effective partial discharge detection system can lead to significant savings on maintenance and prevent power disruptions. Traditional methods rely on hand-crafted features and domain expertise to identify partial discharge patterns in the electrical current. Many data-driven deep learning-based methods have been proposed in recent years to remove these ad hoc feature extraction. However, most of these methods either operate in the time-domain or frequency-domain. Many research approaches have been developed to generate phase-resolved partial discharge (PRPD) patterns from raw PD sensor data. They extract the salient characteristics of these PRPD patterns and provide a visual interpretation system for a comprehensive diagnosis of the defects. These PRPD diagrams suggest a correlation between partial discharge activities occurring in an alternating electrical waveform's positive and negative half-cycles. However, this correlation criterion between half-cycles has been remained unexplored in deep learning-based methods. This work proposes a novel feature-fusion-based Dual-CyCon Net that can utilize all time, frequency, and phase domain features for joint learning in one cohesive framework. Our proposed cycle-consistency loss exploits any relation between an alternating electrical signal's positive and negative half-cycles to calibrate the model's sensitivity. This loss explores cycle-invariant PD-specific features, enabling the model to learn more robust, noise-invariant features for PD detection. A case study of our proposed framework on a public real-world noisy measurement from high-frequency voltage sensors to detect damaged power lines has achieved a state-of-the-art MCC score of 0.8455, demonstrating the effectiveness of joint learning and cycle-consistency loss.
\end{abstract}


%
\IEEEpeerreviewmaketitle

\section{Introduction}
\label{Introduction}
\begin{figure}[!t]
	\centering
	\includegraphics[width=\linewidth]{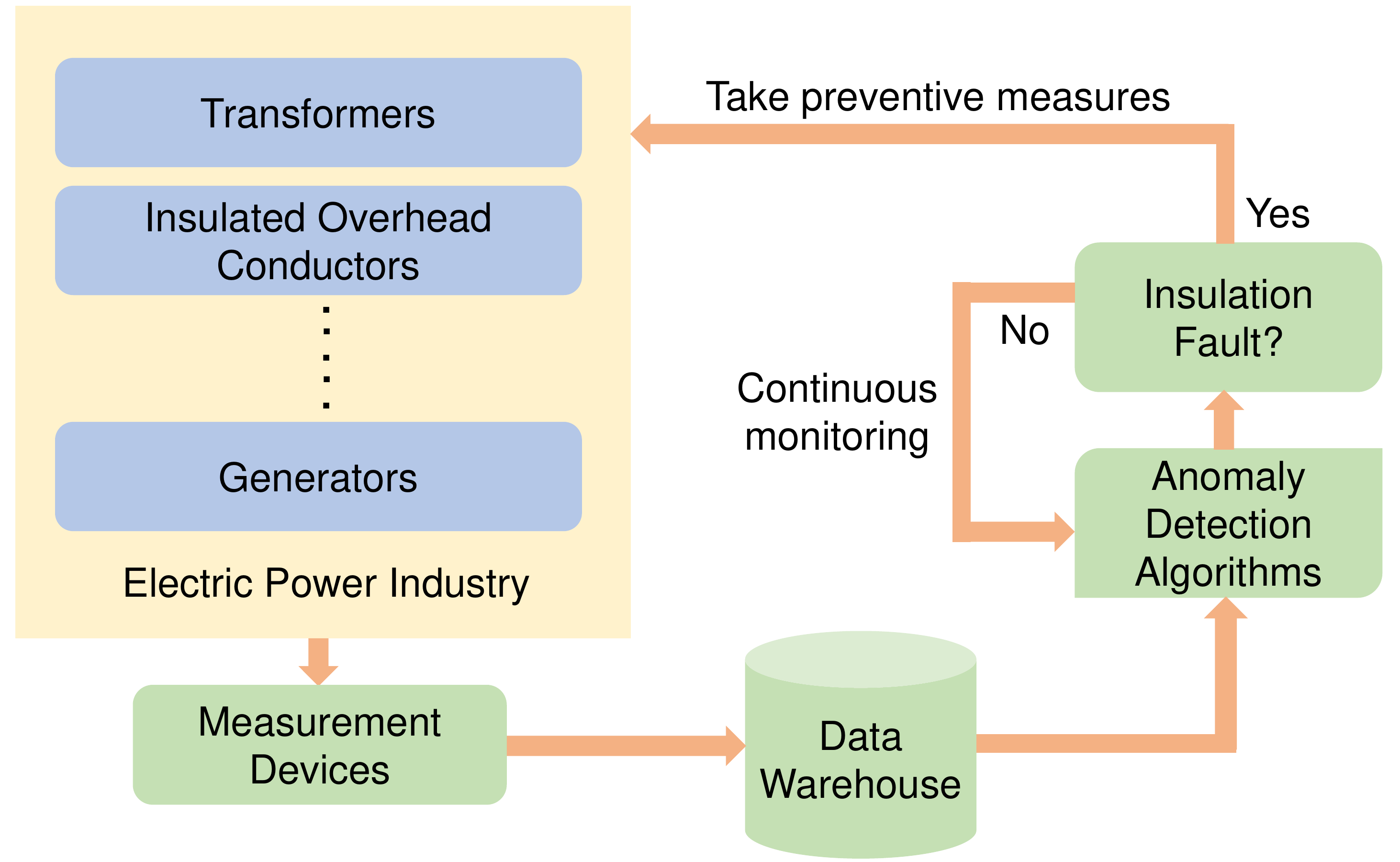}
	\caption{Overview of an automated insulation fault detection scheme in the electric power industry.}
	\label{generalized_preview}
\end{figure}

Modern society infrastructures are built upon the backbone of robust, reliable, and undisrupted power generation and distribution of electricity. A disruption in the power distribution system can cause severe negative impacts in a society's industrial sectors that rely on the continuous power supply. The electric power industry must have high reliability and trustworthy fault detection systems to guarantee the stability and continuity of power distribution operations \cite{7962173}. One of the prominent causes for failure in the electrical equipment is the degradation of the insulation system. Continuous monitoring of the electrical equipment is required for the early identification of insulation failure to prevent the complete breakdown of a system. Intelligent monitoring systems such as computer-aided analysis of insulation failure can lead to significant savings on maintenance and avoid disruptions of power. Automation of industrial fault detections is one of the leading areas with practical applications for recently developed deep learning algorithms. Fig. \ref{generalized_preview} represents a generalized view of an active monitoring system for detecting insulation breakdown.

\begin{figure*}[!t]
	\centering
	\includegraphics[width=\linewidth]{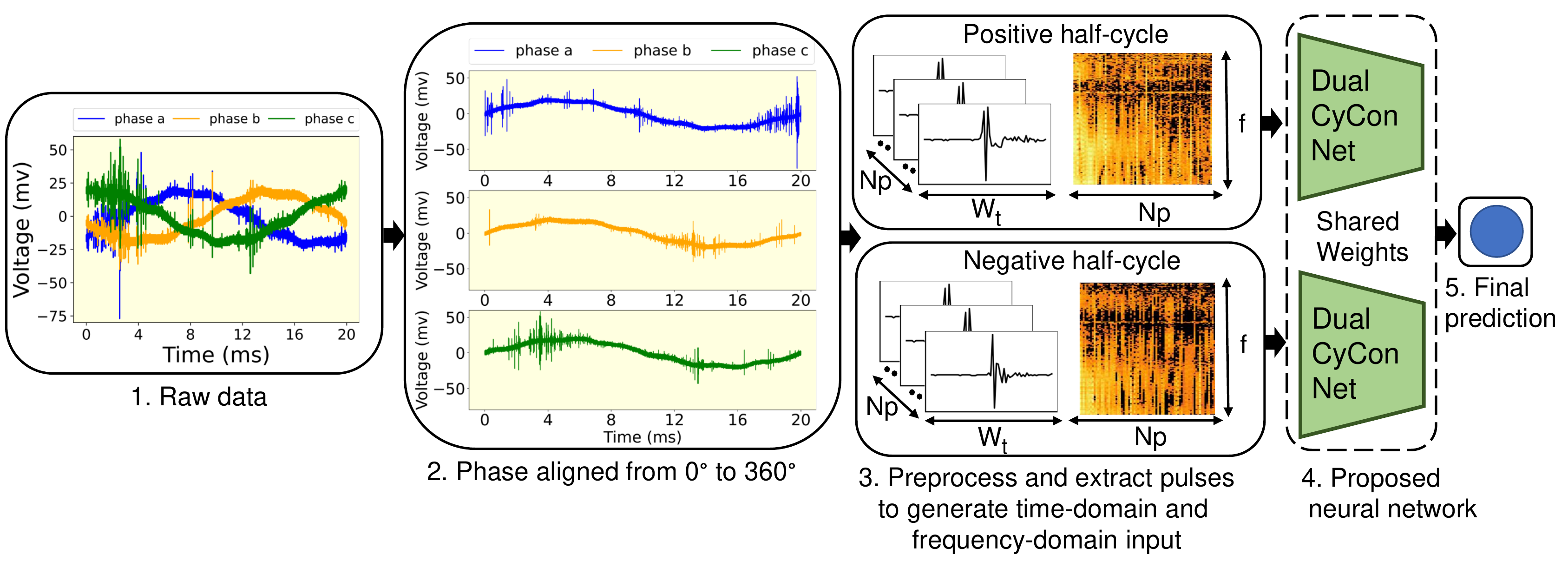}
	\caption{Graphical Abstract: (1) Three-phase raw data over one period of grid frequency. (2) The signals are shifted and rearranged from $0\degree$ to $360\degree$ phase angle. (3) $N_p$ number of pulses with a window of length $W_{t}$ is extracted from positive and negative half-cycles of the three-phase signal with a preprocessing method described in section \ref{Preprocessing}. Spectrograms with $f$ number of frequency bins are generated from these time series corresponding to each peak $N_p$. (4) Time-domain pulse arrays and spectrograms from positive and negative half-cycles are fed to the shared Dual-CyCon Net. (5) The output of the model is considered to label the health status of the powerline.}
	\label{overall_archi}
\end{figure*}

Medium voltage overhead power lines run for hundreds of miles to supply power to cities. These power lines utilize insulated overhead conductors (IOC) in many places around the world \cite{6737537, Dabbak2015SurfaceDC, Pakonen2007DetectionOI}. However, these long distances make it expensive to manually inspect the lines for damage that does not immediately result in a power outage, such as a tree branch hitting the line or an insulator flaw. Standard overcurrent protection devices also can not detect these types of faults \cite{7962192, 1038663}. Although invisible and undetectable to standard protection devices, these modes of damage lead to a phenomenon known as partial discharge (PD) \cite{7422591,7422593,7480674,1678355}. PD is an electrical discharge that does not bridge the electrodes between an insulation system completely. PD occurs across the surface of insulating material where the electric field strength exceeds the breakdown strength of the insulating material \cite{6957620,1430395, Hashmi2010ModelingAE,7422591}. Partial discharges slowly damage the power line by degrading the insulation system of the IOC \cite{7962192,7909221}, so left unrepaired, they will eventually lead to a power outage or start a fire \cite{1038663}. Therefore, it is of utmost importance to continuously assess the condition of electrical equipment to detect these events to take proper preventative actions accordingly. The main obstacle to identifying PD activity lies in detecting extremely short and temporally localized events \cite{4073965} and the presence of the background noise interference \cite{10635_15365}.

A significant amount of research efforts have been carried out for the computer-aided detection of PD pattern detection. Feature engineering-based methods have extracted relevant handcrafted features and utilized classical mechanical models such as support vector machine (SVM) or random forest to identify PD activity through these features \cite{7909221,8861809,article_shang}. Many data-driven deep-learning approaches have been adopted over time to remove ad hoc feature engineering processes \cite{7757816, 8642226, article_Nguyen, doi:10.1063/5.0011998}. Qu et al. \cite{9087854} utilized discrete wavelet transform and long short-term memory (LSTM) for PD pattern detection, while Wang et al. \cite{9183469} proposed a spectrogram feature extraction-based neural network approach for PD detection. Michau et al. \cite{Michau2021InterpretableDO} proposed a framework for PD detection in time series through a temporal convolutional neural network (CNN).

However, to the best of our knowledge, almost every deep learning-based previous research approaches work either in the time domain \cite{8642226, 9087854, Michau2021InterpretableDO} or frequency domain \cite{9183469}. Performances from each research suggest that salient features characterizing PD activity are present in both time and frequency domains. A model architecture that can utilize joint learning by implementing knowledge flow between these domains can further enhance the model's capability to identify the pulses responsible for PD discharge. The phase-resolved partial discharge (PRPD) pattern-based algorithms rely on the generation and analysis of PRPD diagram for detection of PD activity \cite{1430399, 4446759, en14092567, 9215344, Altenburger2002AnalysisOP}. The PRPD diagram in these research works implies a relation between the occurrences of PD activities, the phase angle, and the voltage amplitude of an alternating electrical waveform. This correlation suggests consistent PD activities in positive and negative half-cycles of an alternating electrical waveform at a relatively same phase angle. However, this criterion has been remained unexplored in deep learning-based approaches.

This paper proposes a novel cycle-consistency aware architecture named Dual-CyCon Net along with novel DDAM structured block and cycle-consistency loss. Our proposed dual-domain attention module (DDAM) enables collaborative learning by allowing the flow of knowledge between time and frequency domain branches. We propose a cycle-consistency loss between positive and negative half-cycles of a particular alternating electrical signal to enhance the model's ability to learn more robust, noise-invariant PD detection features. The cycle-consistency loss integrates the capability for exploiting knowledge distillation between these half-cycles. We validate our proposed architecture on a public dataset of noisy real-world measurements from high-frequency voltage sensors \cite{vsb_dataset_manual}. An overview of our proposed framework is presented in Fig. \ref{overall_archi}. The contributions of this paper are summarized as follows:

\begin{itemize}
  \item We present a novel cycle-consistency aware architecture named Dual-CyCon Net that integrates time, frequency, and phase domain characteristics in one cohesive framework.  
  \item Our proposed dual-domain attention module enhances the capability of our model to identify PD patterns by utilizing knowledge from all available domains.
  \item Cycle-consistency loss tweaks the sensitivity of the model to learn more robust and noise invariant features for PD detection by exploiting the correlation of PD-specific features present in both positive and negative half-cycles of a particular alternating electrical signal.
  \item We evaluate our proposed framework on a real-world noisy public dataset. Our proposed Dual-CyCon Net has achieved a state-of-the-art MCC score of 0.8455, proving the efficacy of DDAM and cycle-consistency loss.
\end{itemize}

The rest of the paper is organized as follows. Section \ref{Related work} highlights the related work. Section \ref{Methodology} describes Dual-CyCon Net along with the motivation of our study and preprocessing of the dataset. The training schemes for our proposed method are presented in Section \ref{Training}. The results and analysis of comprehensive experiments and the ablation study are given in Section \ref{Experimental Results And Ablation Study}. Finally, Section \ref{Conclusion} concludes the whole work.

\section{Related work}
\label{Related work}
The issue of PD detection is a subject of many areas of study, including artificial intelligence, signal processing, data analysis, statistic, or applied mathematics \cite{8535878,8535913,8341662,HAN2019474}. Phase-resolved partial discharge (PRPD) based methods take advantage of the property that for some systems, PD always occurs at the same phase angle in the alternating electrical current \cite{1430399,4446759,en14092567,9215344,Altenburger2002AnalysisOP}. These methods implement algorithms to process the raw values from PD sensors to generate the PRPD diagrams. Experts can visually analyze and interpret these diagrams to identify the traces left by PD activity. However, aggregation of pulses over several hundred or thousands of periods and the presence of noise or superimposed pulses pose serious difficulty in analyzing these PRPD diagrams. The limitations of PRPD analysis have inspired researchers to explore statistically feature engineering-based approaches \cite{8712431}.

In 2014, Shang et al. \cite{article_shang} proposed a novel method based on a multi-kernel multi-class relevance vector machine for PD recognition. In 2017, Misak et al. \cite{7909221} proposed a random-forest-based solution optimized with self-organized migrating optimization. They used the univariate denoising method and different layers of complex preprocessing to remove noisy peaks generated by discrete spectral interferences, repetitive pulses interference, and ambient noise. In 2019, Dong et al. \cite{8861809} developed a unique method based on Seasonal and Trend decomposition using Loess (STL) and utilized Support Vector Machine (SVM) to recognize PD activities on insulated overhead conductors. However, extracting relevant features is time-consuming and identifying characteristics such as different types of pertinent entropy to PD requires years of domain expertise.

Many data-driven deep learning approaches have been applied in recent studies to address the feature extraction challenges mentioned above. Li et al. \cite{7757816} proposed a CNN based deep architecture, which is established to extrapolate new features automatically to realize ultra-high frequency signals recognition in gas-insulated switchgear (GIS). Adam et al. \cite{8642226} applied LSTM to identify different types of PD activity in insulated cables, using single PD impulses as input data. In 2018, Nguyen et al. \cite{article_Nguyen} proposed an approach for detecting PD patterns in GIS using LSTM and recurrent neural networks. Khan et al. \cite{article_Khan} utilized a time-domain CNN for end-to-end partial discharge pattern detection in power cables. Wang et al. \cite{doi:10.1063/5.0011998} applied a CNN along with PRPD criteria to effectively and efficiently distinguish the types of GIS PD. In 2020, Qu et al. \cite{9087854} used discrete wavelet transformation to decompose the signal and extract features with different resolutions to feed them to an LSTM model. Dong et al. \cite{Dong2019PartialDD}, they employed LSTM to identify PD activity by extracting features from the STL residual. Wang et al. \cite{9183469} utilized spectrograms generated from the time-series data and fed them to CNN for PD detection. Michau et al. \cite{Michau2021InterpretableDO} developed a no-feature-engineering end-to-end learning framework for PD identification.  
 
\section{Methodology}
\label{Methodology}

\subsection{Motivation}
\label{Motivation}
Over the last decade, deep learning-based methods for PD detection have worked either in the time domain or frequency domain. Classical approaches have extracted features from time-series data and used them for PD classification. However, to the best of our knowledge, no other approaches have tried to implement algorithms that can utilize joint learning between these domains. A single model that can incorporate knowledge from all available domains such as time series, frequency bins, and phase angle property, will lead to a more robust detection technique. Our proposed DDAM block utilizes joint learning to enable this flow of knowledge between different domains to reinforce the efficiency of the learning capability of the model. This block uses the peak attention vector to adaptively recalibrate peak-wise feature responses by explicitly modeling interdependencies between peaks. Apart from the joint learning, PRPD inspired consistent PD activity between positive and negative half-cycles of an alternating electrical waveform has been remained unexplored in deep-learning-based approaches. Cycle-consistency loss is motivated by the requirement that the high-level features extracted from a particular signal's positive and negative half-cycles be consistent. If the distribution of high-level features differs considerably, the model has not yet learned how to attenuate temporal ambient noise introduced in the signal's positive or negative half-cycle. PD traces left in a particular signal's positive and negative half-cycles come from the same physical fault in the insulation. The cycle-consistency loss allows the model to learn more PD-specific and noise-invariant high-level features by exploiting knowledge distillation between half-cycles.

\subsection{VSB ENET dataset}
\label{VSB ENET Dataset}
We evaluate our proposed framework on the VSB dataset, generated and released by the Technical University of Ostrava \cite{vsb_dataset_manual, 6957620}. The objective of this dataset is to detect damaged three-phase, medium-voltage overhead power lines by identifying PD patterns in the observed signals. The dataset contains $2904$ measurements, each containing three-phase waveforms, totalling $8712$ individual samples. Each sample is associated with a label indicating whether the power line insulation was damaged at the time of recording. However, no additional information regarding PD types, shapes, or location is provided. Out of $8712$ samples, $575$ samples are labelled as damaged power lines. The electric voltage waveform is recorded over one period of the grid frequency ($50$Hz) for all three phases simultaneously for each measurement. Each of the samples contains $800,000$ values which are recorded with a sampling frequency of $40$MHz. The $800,000$ sample points constitute one whole cycle containing both positive and negative half-cycles. 

\subsection{Preprocessing}
\label{Preprocessing}
\begin{figure}[!t]
     \centering
     \begin{subfigure}[t]{0.48\textwidth}
         \centering
         \includegraphics[width=\textwidth,height=0.26\textwidth]{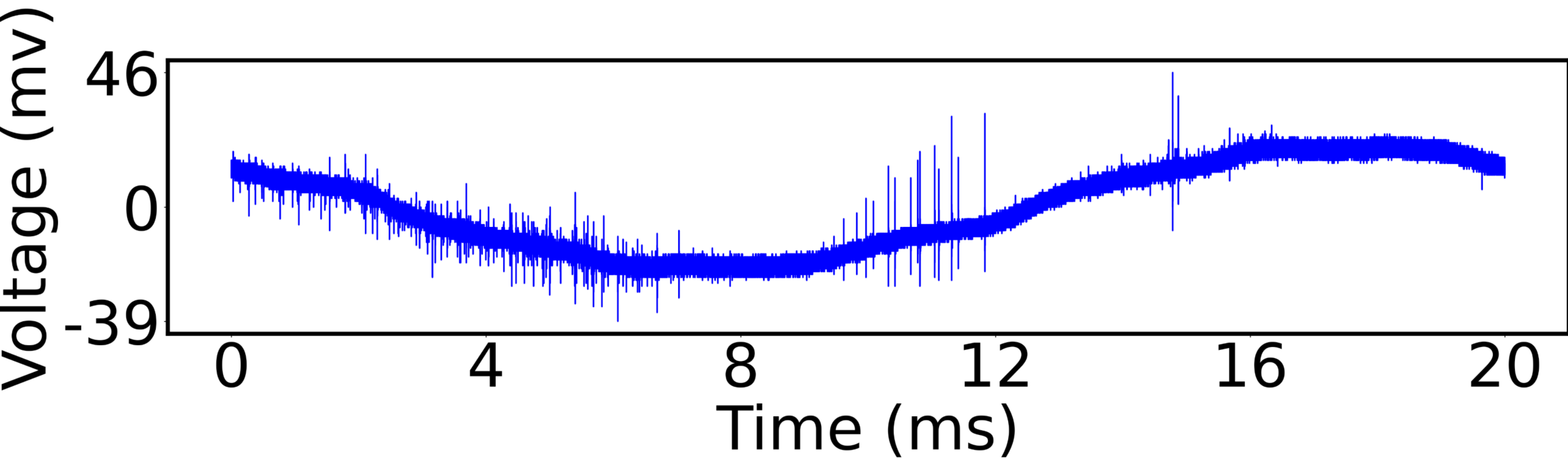}
         \caption{Raw signal.}
         \label{prepro:a}
     \end{subfigure}
     \begin{subfigure}[t]{0.48\textwidth}
         \centering
         \includegraphics[width=\textwidth,height=0.26\textwidth]{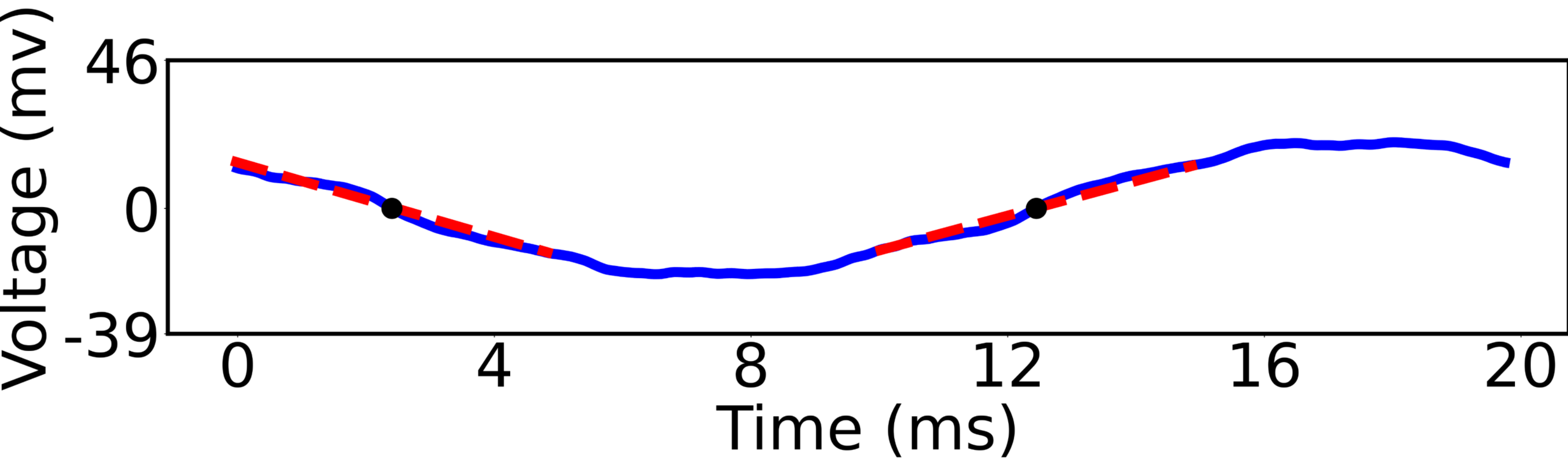}
         \caption{Zero-crossing points detection.}
         \label{prepro:a2}
     \end{subfigure}
     \begin{subfigure}[t]{0.48\textwidth}
         \centering
         \includegraphics[width=\textwidth,height=0.26\textwidth]{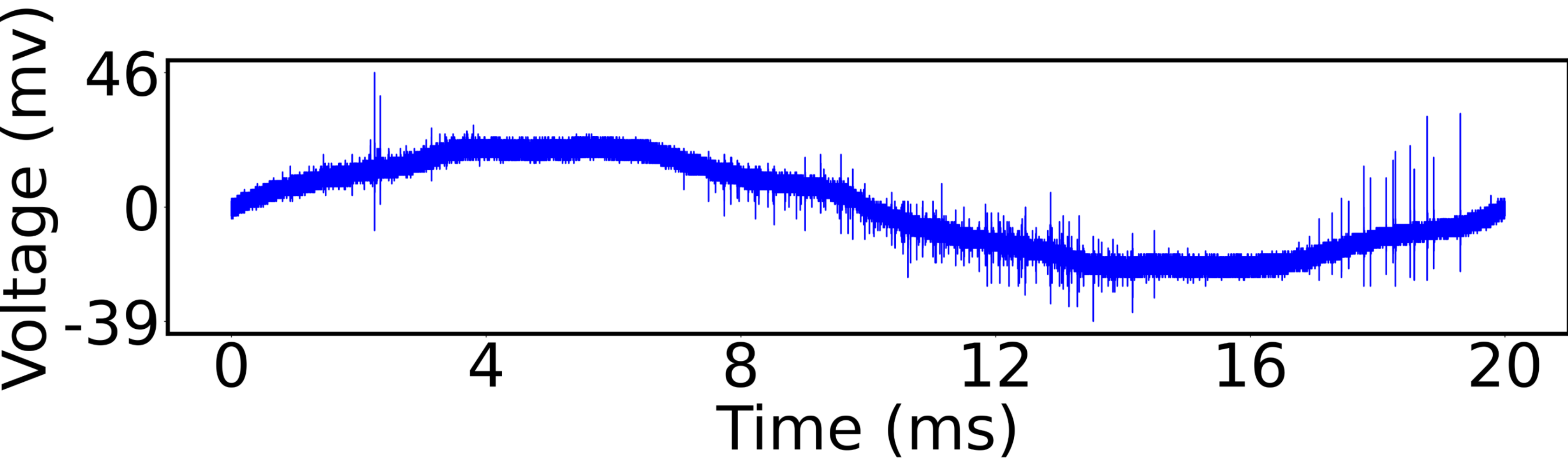}
         \caption{Phase aligned signal.}
         \label{prepro:b}
     \end{subfigure}
     \begin{subfigure}[t]{0.48\textwidth}
         \centering
         \includegraphics[width=\textwidth,height=0.26\textwidth]{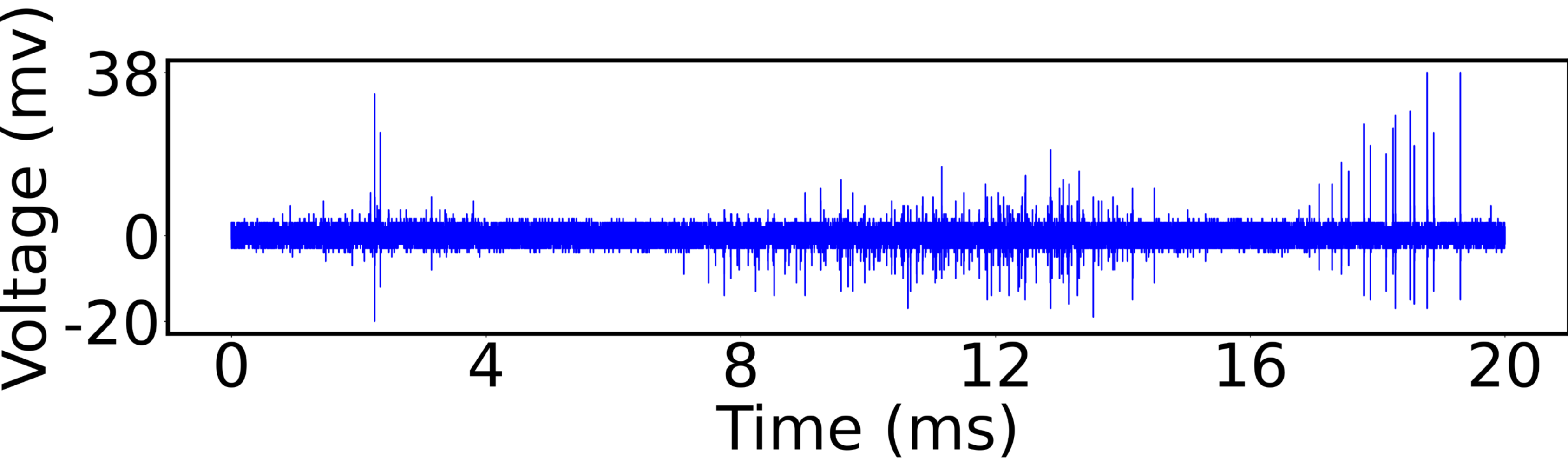}
         \caption{High-pass filtered signal.}
         \label{prepro:c}
     \end{subfigure}
     \begin{subfigure}[t]{0.48\textwidth}
         \centering
         \includegraphics[width=\textwidth,height=0.26\textwidth]{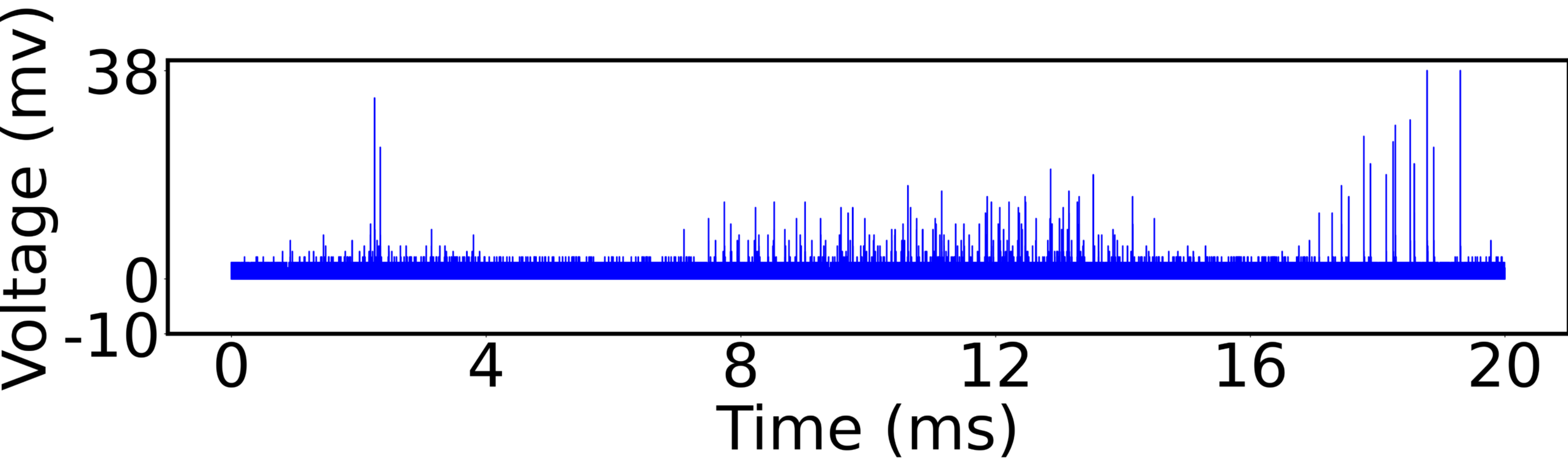}
         \caption{Absolute high-pass filtered signal.}
         \label{prepro:d}
     \end{subfigure}
     \begin{subfigure}[t]{0.48\textwidth}
         \centering
         \includegraphics[width=\textwidth,height=0.26\textwidth]{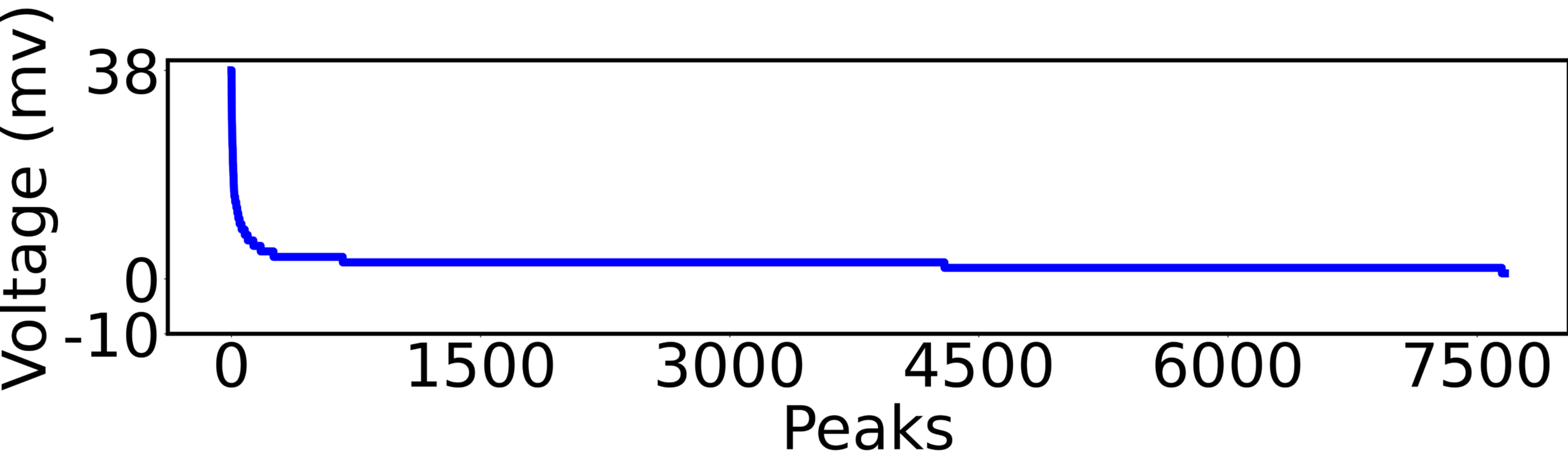}
         \caption{Peak heights (sorted in descending order).}
         \label{prepro:e}
     \end{subfigure}
     \begin{subfigure}[t]{0.48\textwidth}
         \centering
         \includegraphics[width=\textwidth,height=0.26\textwidth]{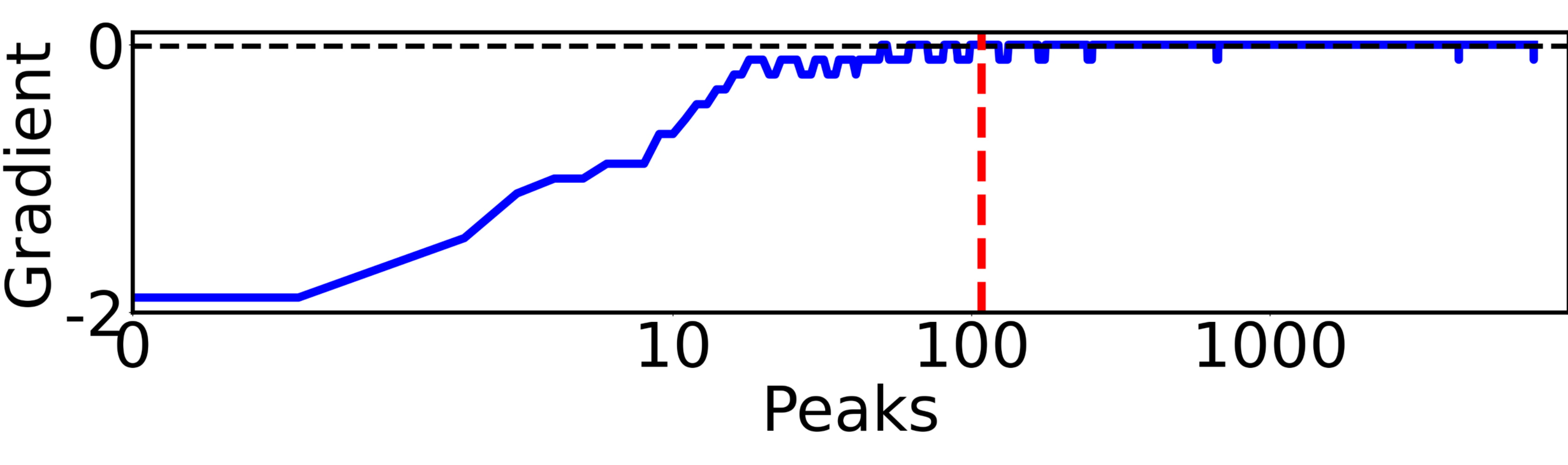}
         \caption{Peak filtering.}
         \label{prepro:f}
     \end{subfigure}
     \begin{subfigure}[t]{0.48\textwidth}
         \centering
         \includegraphics[width=\textwidth,height=0.26\textwidth]{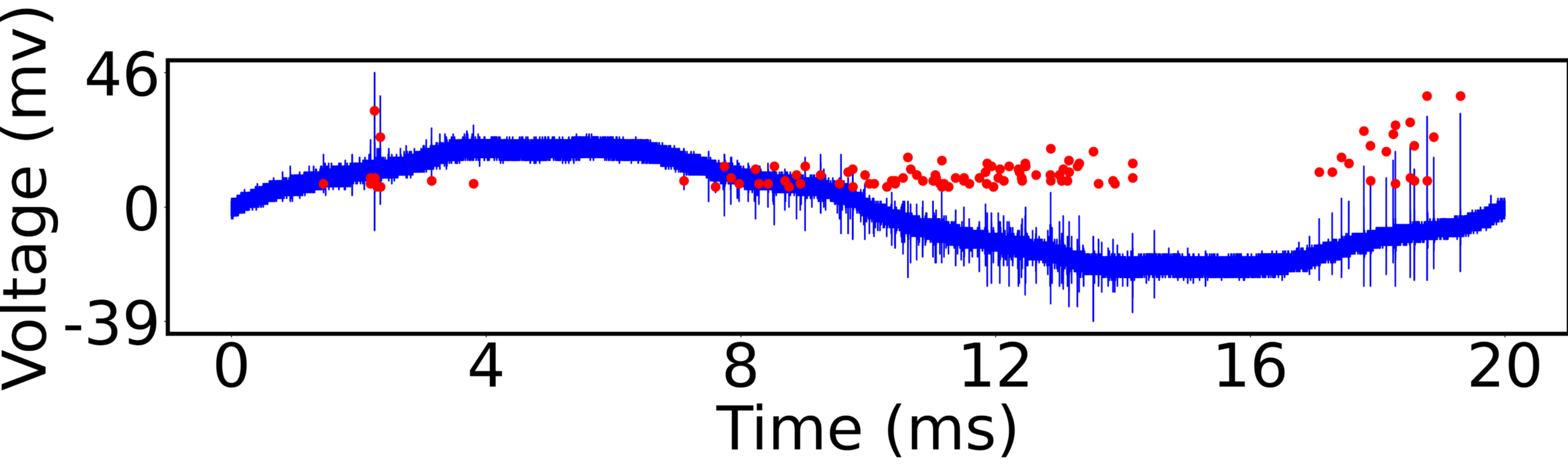}
         \caption{Signal with filtered peaks.}
         \label{prepro:g}
     \end{subfigure}
        \caption{Visualization of the signal from each step of the preprocessing algorithm.}
        \label{preprocessing_fig}
\end{figure}

Only a few PD pulses can occur per period of the current utility frequency \cite{4073965}. Inspired by the preprocessing steps in \cite{Michau2021InterpretableDO, 9183469, vsb_kaggle_preprocess}, we identify and extract the pulses from a particular signal. Let our signal from a phase of a particular three-phase measurement is denoted by $\mathbf{X} \in \mathbb{R}^{N}$. $N$ represents the number of sample points for a signal. For our dataset, $N=800000$. $\mathbf{X}$ is shown in Fig. \ref{prepro:a}. First, we shift and rearrange $\mathbf{X}$ from the phase angle $0\degree$ to $360\degree$ degrees. That enables us to separate that particular signal's positive and negative half-cycles. We apply a $10000$ sample points moving average filter to smooth out the signal \cite{principles_maf}. Afterward, we detect the zero-crossing sample points and calculate the gradients of them with respect to neighboring sample points. Based on the gradient sign, we identify sample points with phase angles $0\degree$ and $180\degree$. Positive gradient denotes sample point with $0\degree$ phase angle, while negative gradient denotes sample point with $180\degree$ phase angle. This is shown in Fig. \ref{prepro:a2}. By shifting and aligning, we get phase-resolved signal $\mathbf{X}_p$, which starts at phase angle $0\degree$ and ends in $360\degree$. $\mathbf{X}_{p}$ is shown in Fig. \ref{prepro:b}. PDs are due to insulation failures and typically occur at particular voltage changes. They are visible in a much higher frequency band than the power grid frequency $f_{ut}$. So we require to remove the low frequencies \cite{Michau2021InterpretableDO}. We flatten the signal to remove the lower frequencies by,

\begin{align}
& \mathbf{z}_i = \frac{\mathbf{z}_{i-1}*(\alpha-\beta)}{\alpha} + \frac{\beta (\mathbf{X}_{p})_{i}}{\alpha} \\
& (\mathbf{X}_{hp})_{i} = (\mathbf{X}_{p})_{i} - \mathbf{z}_i
\end{align}

where, $i \in \{1,2,....,N\}$, $\mathbf{z}_{1}=\mathbf{X}_{1}$, $\alpha$=$100$, and $\beta$=$1$. The $\mathbf{X}_{hp}$ signal is shown in Fig. \ref{prepro:c}. Then, we take the absolute value of each sample point of the signal $\mathbf{X}_{hp}$ and get $\mathbf{X}_{d}$ which is shown in Fig. \ref{prepro:d}. Based on simple maximum filter, we extracted $N_{p1}$ number of pulses and their corresponding heights $h_{p1}$ from $\mathbf{X}_{d}$ \cite{Michau2021InterpretableDO}. To remove noisy peaks from them, we sorted the peaks $N_{p1}$ according to heights $h_{p1}$ and generate gradients $\mathbf{g}_{p1}$ of the heights. After that, we smooth out the gradient $\mathbf{g}_{p1}$ by linear convolution with a vector $\mathbf{v}$ of length $L$ for utilizing knee point detection,

\begin{align}
& \mathbf{v}_{j} = \frac{j}{L}, \ j \in \{1,2,...,L\}, L=9 \\
& (\mathbf{g}_{p1})_i = h_{i+1} - h_{i}, i \in \{1,....,N_{p1}-1\} \\
& (\mathbf{g}_{p1})_{smooth} = \mathbf{g}_{p1} * \mathbf{v}
\end{align}

The sorted peaks according to their heights are shown in Fig. \ref{prepro:e}. The gradient is shown in the semi-logarithm scale in Fig. \ref{prepro:f}. Next, based on knee point detection, we figure out where the gradient flattens. The flatten region refers to the constant noise that is intrinsic to the PD measurement sensor. This region does not contain any valuable information as it almost exists at all phase angles of all signals. For that, we only keep pulses that are before the flatten. This processing will reduce the impact of background noises \cite{Dong2019PartialDD}. We then plotted these peaks on the original signal in Fig. \ref{prepro:g}.

\begin{figure}[!t]
	\includegraphics[width=\linewidth]{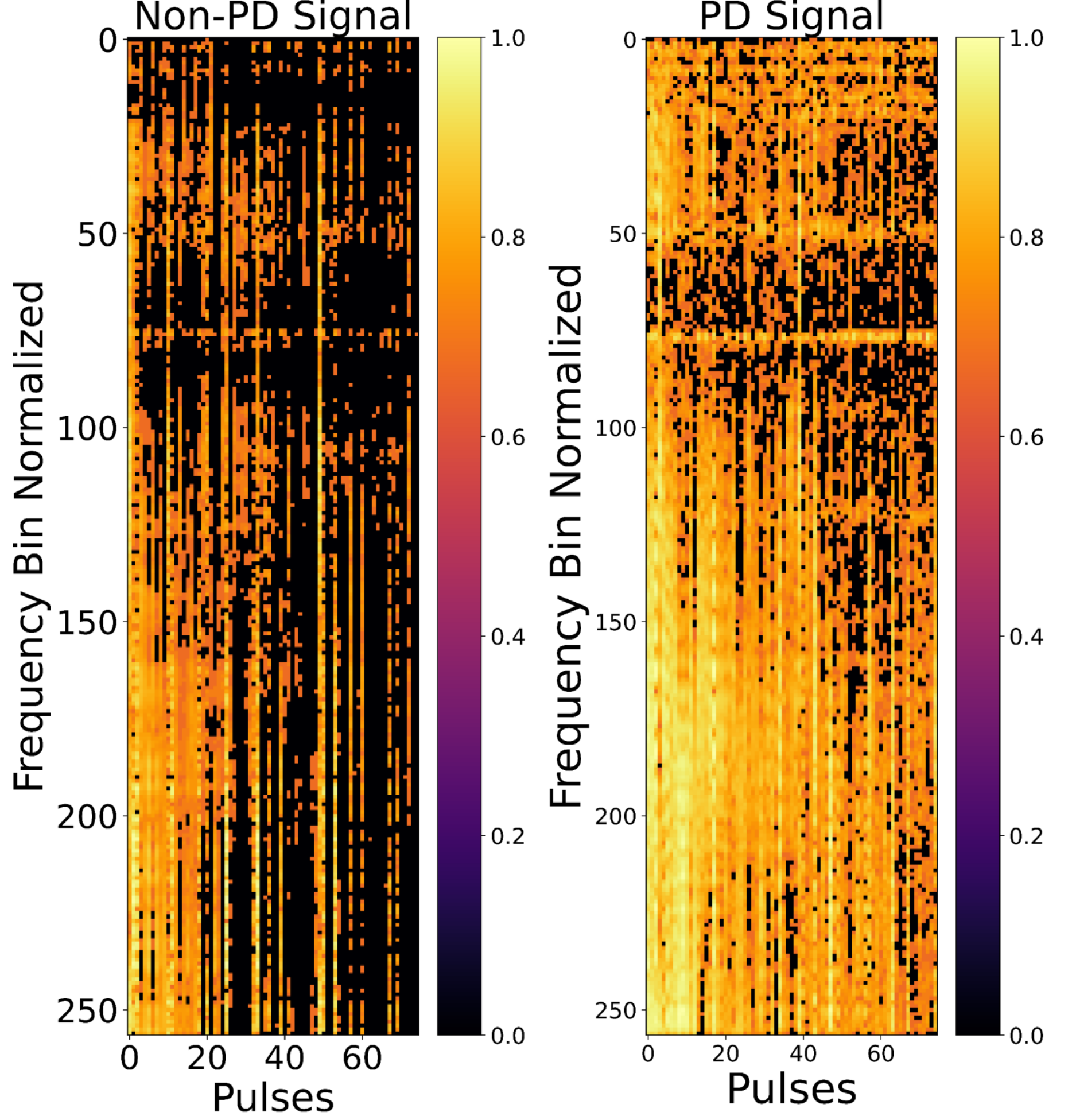}
    \caption{Example of log spectrograms. Left one demonstrates the log spectrogram of a non-PD signal. The right one shows the log spectrogram of a PD signal.}
    \label{spectrogram_example_fig}
\end{figure}

Let $N_{p}^{ap}$, $N_{p}^{bp}$, $N_{p}^{cp}$ are the number of peaks found from the positive half-cycles of three-phase conductors $A$, $B$, and $C$ for a particular measurement. Peaks that are found before phase angle $180\degree$ are considered as positive half-cycle peaks, while the rest of the peaks are considered as negative half-cycle peaks. These positive half-cycle peaks from three-phase signals are concatenated together and sorted in descending order according to the absolute value of their heights. Afterward, we select the first $N_{p}$ number of peaks from these sorted peaks. Similarly, we extract $N_{p}$ number of peaks from the concatenated and sorted $N_{p}^{an}$, $N_{p}^{bn}$, $N_{p}^{cn}$ peaks for the negative half-cycle. Thus, we get two sets of $N_{p}$ number of peaks, one for positive half-cycle, another for negative half-cycle. Then, we extract data from the signal $\mathbf{X}_{hp}$ utilizing these local maximal values using a window size of $w_{t}$ and $w_{f}$. That is, if the $i^{th}$ local maximum is localized at timestamp $t_{i}$ and if we use a window of size $w$, we extract the interval $[t_{i}-\frac{w}{2}, t_{i}+\frac{w}{2}]$. The collection of pulses is, therefore, two 2D array of shape $N_{p} \times w_{t}$ and $N_{p} \times w_{f}$ separately for positive and negative half-cycles. For our case, $N_{p}$=257, $w_{t}$=128, $w_{f}$=512, which is determined through grid search in cross-validation. If a fewer number of peaks are detected in the pulse extraction step, the rest of the peaks are considered as zero-padded. This array of dimension $N_{p} \times w_{t}$, from both positive and negative half-cycle, is considered as time-domain input for our framework. For frequency-domain input, we converted $N_{p} \times w_{f}$ array from both positive and negative half-cycles to spectrograms by,

\begin{multline}
\mathbf{X}(i,k) = \sum_{m=0}^{N_{w}-1} \mathbf{x}[iN_{h}+m]\mathbf{\omega}[m]exp(\frac{-j2\pi km}{n}), \\ 0 \leq k \leq n-1
\end{multline}

where $\mathbf{X}(i, k)$ is the short term fourier transform (STFT) of subsequence i of $\mathbf{x}$ with frequency $k$. $\omega$ is the window function with window size $N_{w}$. $N_{h}$ is the hop size. The hop size is defined as $N_{h}=N_{w}-N_{o}$, where $N_{o}$ is overlap between consecutive subsequences. We select the hanning window as window function. We apply STFT across the peak axis $N_{p}$. So, $N_{o}$ and $N_{h}$ is zero in our case. The window size $N_{w}$ is set to $w_{f}$. The resultant spectrogram is $\mathbf{S} \in \mathbb{R}^{N_{p} \times (w_{f}/2)+1}$. Now, the log-spectrogram $\mathbf{S}_{log}$ can be obtained by $\mathbf{S}_{log}(i, k) = log(|\mathbf{X}(i, k)|^{2})$. An example showcasing spectrograms of a PD and non-PD signal is shown in Fig. \ref{spectrogram_example_fig}. The time-domain and frequency-domain inputs are normalized respectively to $[-1,1]$ and $[0,1]$ range before feeding to the neural network.

\subsection{Dual-CyCon Net}

\begin{figure*}[!t]
	\centering
	\includegraphics[width=\linewidth]{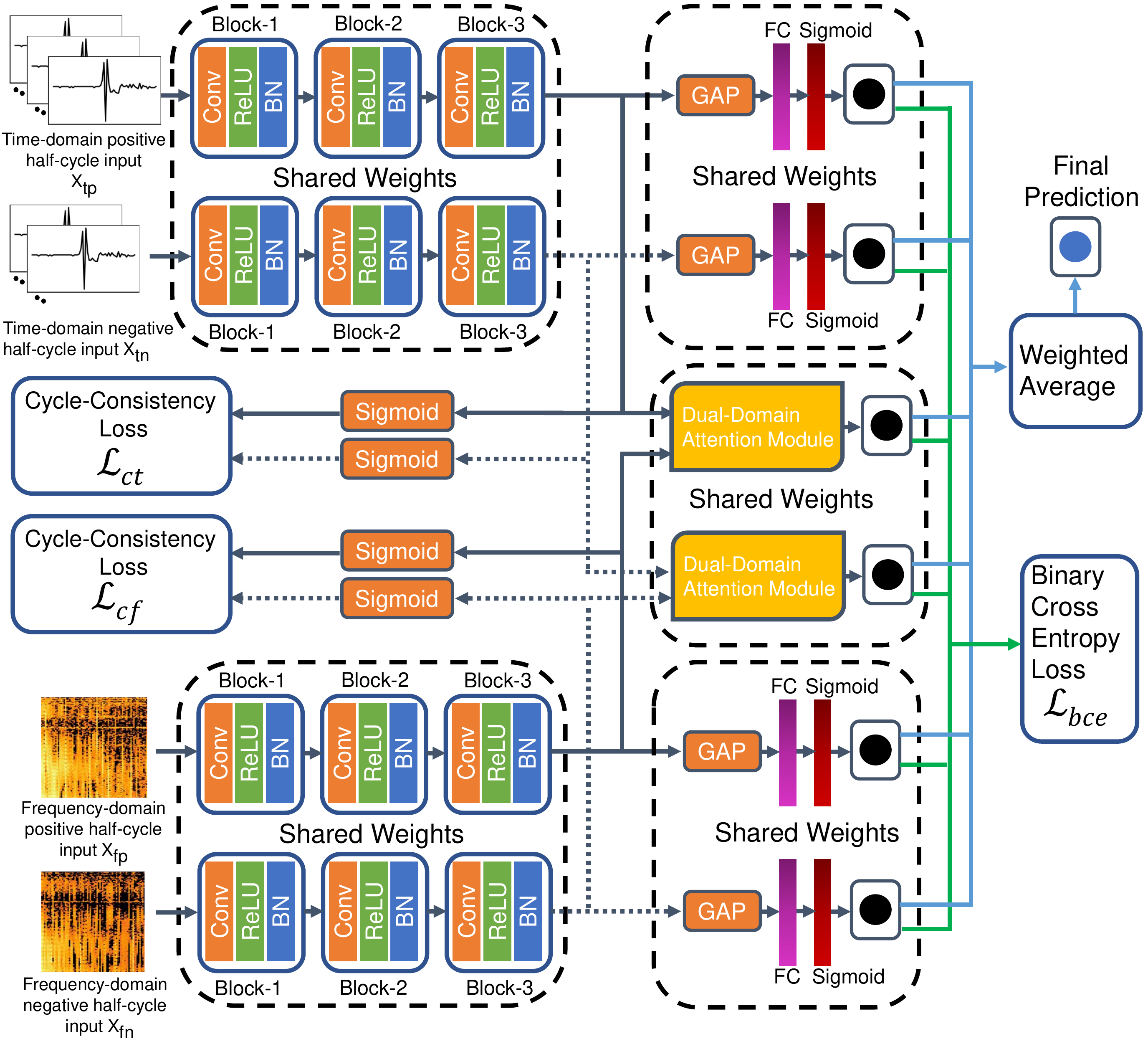}
	\begin{tabular}{cccc}
    * GAP: Global average pooling & FC: Fully connected & BN: Batch normalization & ReLU: Rectified linear unit
    \end{tabular}
	\caption{The architecture of the proposed Dual-CyCon Net. Time-domain inputs for both positive and negative half-cycles are passed to the shared time-domain branch of the framework. Similarly, frequency-domain inputs for both positive and negative half-cycles are fed to the shared frequency-domain branch of the framework. The cycle-consistency loss works on the output from block-3 of these shared branches. The output of block-3 from both time-domain and frequency-domain branches for a particular half-cycle are fed to the DDAM block for joint learning. Each feature space from block-3 of time-domain and frequency-domain is passed through a global average pooling layer, a shared fully connected layer, and a sigmoid layer. Dual-CyCon Net makes predictions based on the weighted average of the output from these fully connected layers and the output from the DDAM block. The classification loss works on all of these outputs.}
	\label{dualcycon_archi}
\end{figure*}

The proposed framework is illustrated in Fig. \ref{dualcycon_archi}. The neural network branches for frequency and time domain are made up of an identical configuration. Both time and frequency domain branches are comprised of three blocks. Each block contains one 2d convolutional layer, followed by one Rectified Linear Unit (ReLU) and a batch normalization layer \cite{pmlr-v37-ioffe15}. The number of blocks, kernel number, and kernel size for these convolutional filters were inferred from a grid search with 5-fold stratified cross-validation. The 2d convolutional layer in all three blocks has a kernel filter of size $7\times7$ with stride $2$ and $0$ padding. They have $8,16,32$ filters respectively for block-1, block-2, and block-3. Let $\mathbf{X}_{tp}$, $\mathbf{X}_{tn}$ $\in$ $\mathbb{R}^{1 \times W_{t} \times N_{P}}$ denote the time-domain input respectively for positive and negative half-cycle, where $W_{t}$ and $N_{P}$ represents respectively the window length and number of peaks taken. Similarly, let $\mathbf{X}_{fp}$, $\mathbf{X}_{fn}$ $\in$ $\mathbb{R}^{1 \times F \times N_{P}}$ denote the frequency-domain input respectively for positive and negative half-cycle, where $F$, $N_{P}$ represents respectively the number of frequency bins and peaks taken.   

\begin{figure*}[!t]
	\centering
	\includegraphics[width=\linewidth]{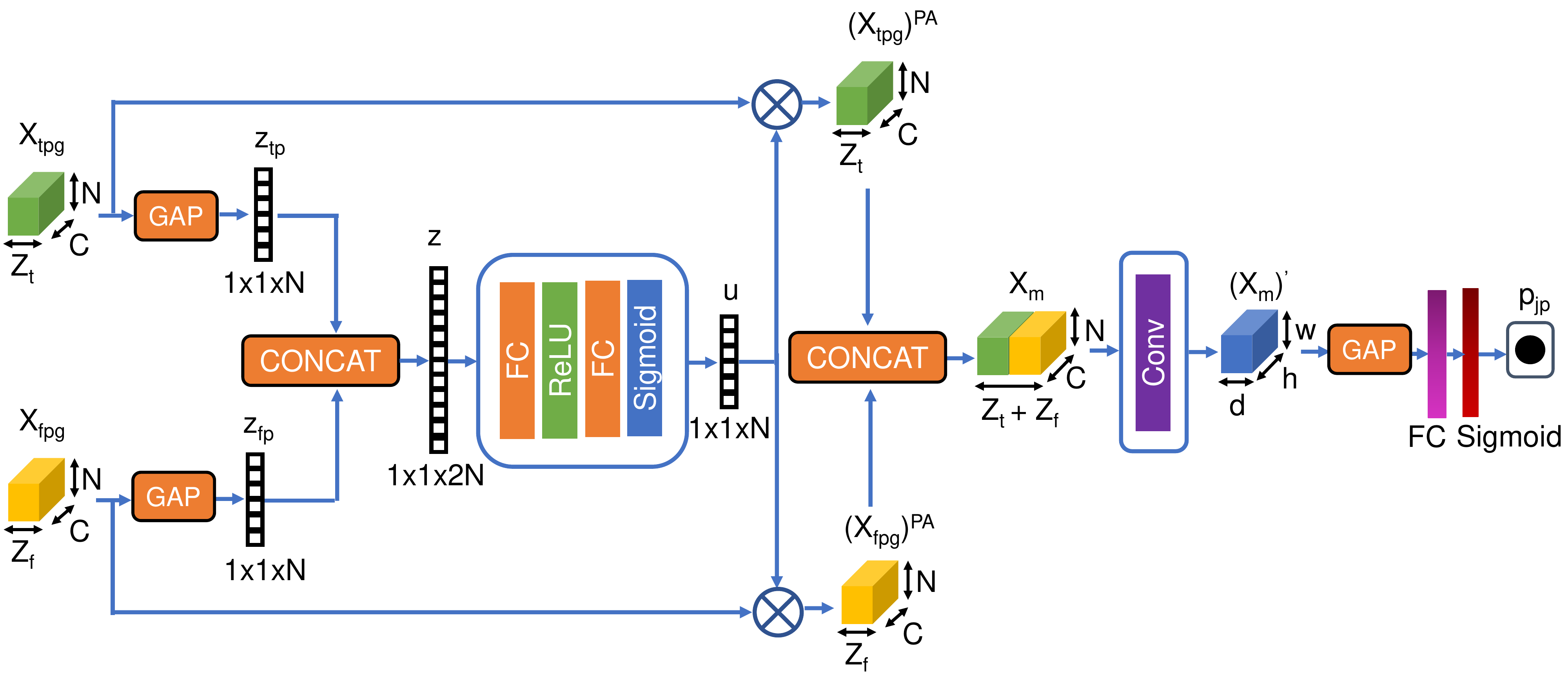}
	\begin{tabular}{cccc}
    * GAP: Global average pooling & FC: Fully connected & CONCAT: Concatenation & BN: Batch normalization
    \end{tabular}
	\caption{The architecture of the proposed DDAM block operating on the positive half-cycle inputs of time and frequency domain. DDAM takes the feature spaces $X_{tpg}$, $X_{fpg}$ respectively from block-3 of the time-domain and frequency-domain branch. It generates peak attention vector $u$ from these feature spaces by utilizing a squeeze and excitation network. Then, it multiplies this attention vector with the feature spaces elementwise in the peak axis to apply the peak attention mechanism. Afterward, these feature spaces are concatenated together and passed through a convolutional layer. This convolutional layer learns salient features characterizing PD activity by using knowledge from both domains.}
	\label{ddam_fig}
\end{figure*}

\subsubsection*{Cycle-Consistency loss}
\hfill \par
The time-domain input arrays $\mathbf{X}_{tp}$ and $\mathbf{X}_{tn}$ are passed through the shared time-domain branch of the proposed framework. Let the feature space from block-3 of the time-domain branch is $\mathbf{X}_{tpg}$ and $\mathbf{X}_{tng}$ respectively for the positive and negative half-cycle. Let $\mathbf{X}_{tpg}$, $\mathbf{X}_{tng}$ $\in$ $\mathbb{R}^{C \times Z_{t} \times N}$, where C is the channel number, $Z_{t} \times N$ is the feature map size. $Z_{t}$ corresponds to high-level time feature dimension, and $N$ is the high-level peak feature dimension. These feature maps are passed through a sigmoid layer and we get probability distributions $\mathbf{p}^{tp},\mathbf{q}^{tn} \in \mathbb{R}^{C \times Z_{t} \times N}$ by,

\begin{align}
& \mathbf{p}^{tp}(i,j,k) = \frac{1}{1 + \exp{\Big(-\mathbf{X}_{tpg}(i,j,k)\Big)}} \\
& \mathbf{q}^{tn}(i,j,k) = \frac{1}{1 + \exp{\Big(-\mathbf{X}_{tng}(i,j,k)\Big)}}
\end{align}

where, $i \in \{1,2,...,C\},\ j \in \{1,2,...,Z_{t}\},\ and \ k \in \{1,2,...,N\}$. Afterward, we calculate cycle-consistency loss between $\mathbf{p}^{tp}$ and $\mathbf{q}^{tn}$ which is a bidirectional Kullback-Leibler divergence loss. We get the cycle-consistency loss $\mathcal{L}_{ct}$ for the time-domain branch by,

\begin{multline}
\mathcal{L}_{ct} = \frac{1}{C \times Z_{t} \times N}\sum_{k=1}^{N}\sum_{j=1}^{Z_{t}}\sum_{i=1}^{C}\mathbf{p}^{tp}_{i,j,k}\log\Big(\frac{\mathbf{p}^{tp}_{i,j,k}}{\mathbf{q}^{tn}_{i,j,k}}\Big) + \\ \frac{1}{C \times Z_{t} \times N}\sum_{k=1}^{N}\sum_{j=1}^{Z_{t}}\sum_{i=1}^{C}\mathbf{q}^{tn}_{i,j,k}\log\Big(\frac{\mathbf{q}^{tn}_{i,j,k}}{\mathbf{p}^{tp}_{i,j,k}}\Big)
\end{multline}

This loss quantifies how much one probability distribution differs from another probability distribution. By minimizing this loss, our model learns to identify high-level PD features that are more noise-invariant. Similarly, the spectrograms inputs $\mathbf{X}_{fp}$ and $\mathbf{X}_{fn}$ are passed through the shared frequency-domain branch of the proposed framework. Let the feature space from block-3 of the frequency-domain branch is $\mathbf{X}_{fpg}$ and $\mathbf{X}_{fng}$ respectively for positive and negative half-cycle. Let $\mathbf{X}_{fpg}$, $\mathbf{X}_{fng}$ $\in$ $\mathbb{R}^{C \times Z_{f} \times N}$, where C is the channel number, $Z_{f} \times N$ is the feature map size. $Z_{f}$ corresponds to high-level frequency bin features dimension and $N$ is the high-level peak features dimension. These feature maps are passed through a sigmoid layer and we get probability distributions $\mathbf{p}^{fp},\mathbf{q}^{fn} \in \mathbb{R}^{C \times Z_{f} \times N}$ by,

\begin{align}
& \mathbf{p}^{fp}(i,j,k) = \frac{1}{1 + \exp{\Big(-\mathbf{X}_{fpg}(i,j,k)\Big)}} \\
& \mathbf{q}^{fn}(i,j,k) = \frac{1}{1 + \exp{\Big(-\mathbf{X}_{fng}(i,j,k)\Big)}}
\end{align}

where, $i \in \{1,2,...,C\},\ j \in \{1,2,...,Z_{f}\},\ and \ k \in \{1,2,...,N\}$. We get the cycle-consistency loss $\mathcal{L}_{cf}$ for the frequency-domain branch by,

\begin{multline}
\mathcal{L}_{cf} = \frac{1}{C \times Z_{f} \times N}\sum_{k=1}^{N}\sum_{j=1}^{Z_{f}}\sum_{i=1}^{C}\mathbf{p}^{fp}_{i,j,k}\log\Big(\frac{\mathbf{p}^{fp}_{i,j,k}}{\mathbf{q}^{fn}_{i,j,k}}\Big) + \\ \frac{1}{C \times Z_{f} \times N}\sum_{k=1}^{N}\sum_{j=1}^{Z_{f}}\sum_{i=1}^{C}\mathbf{q}^{fn}_{i,j,k}\log\Big(\frac{\mathbf{q}^{fn}_{i,j,k}}{\mathbf{p}^{fp}_{i,j,k}}\Big)
\end{multline}

Finally, we get the total cycle-consistency loss by,
\begin{align}
\mathcal{L}_{c} = \mathcal{L}_{ct} + \mathcal{L}_{cf} 
\end{align}

\subsubsection*{Dual Domain Attention Module (DDAM)}
\hfill \par

Our proposed DDAM block takes the feature spaces of block-3 from both the time and frequency domain for a particular half-cycle as its input. Let the feature space from block-3 for time-domain and frequency-domain is $\mathbf{X}_{tpg} \in \mathbb{R}^{C \times Z_{t} \times N}$ and $\mathbf{X}_{fpg}$ $\in$ $\mathbb{R}^{C \times Z_{f} \times N}$ for positive half-cycle of a particular signal. Here C is the channel number, $Z_{t}$, $Z_{f}$, and $N$ represents the dimension of the high-level time-domain features, frequency-domain features, and peak features, respectively.

The architecture of the DDAM block is illustrated in Fig. \ref{ddam_fig}. To give attention to high-level peak features, we adopt the “Squeeze-and-Excitation” operator \cite{8578843} as the peak-wise attention module, which is composed of a global average pooling layer and two consecutive fully connected (FC) layers. The squeeze operation is conducted by applying the global average pooling to the feature map $\mathbf{X}_{tpg}$ and $\mathbf{X}_{fpg}$, resulting in a vector, $\mathbf{z}_{tp},\mathbf{z}_{fp} \in \mathbb{R}^{N}$.

\begin{align}
& \mathbf{z}_{tp}(k) = \frac{1}{C \times Z_{t}}\sum_{j=1}^{Z_{t}}\sum_{i=1}^{C}\mathbf{X}_{tpg}(i,j,k) \\
& \mathbf{z}_{fp}(k) = \frac{1}{C \times Z_{f}}\sum_{j=1}^{Z_{f}}\sum_{i=1}^{C}\mathbf{X}_{fpg}(i,j,k)
\end{align}

Here, $k \in \{1,2,...,N\}$. Then, we concatenate the $\mathbf{z}_{tp}$, $\mathbf{z}_{fp}$ vectors and get a vector $\mathbf{z} \in \mathbb{R}^{2N}$. The excitation process is formulated into two FC layers, which use ReLU and sigmoid activation. In the first FC layer, the number of neurons is set to $\alpha N$, where $\alpha$ is the dimension reduction ratio and is empirically set to $1/4$ \cite{8578843}. In the second FC layer, the number of neurons is set to $N$. Defining the excitation vector as $\mathbf{u}$, the excitation process can be formally expressed as,

\begin{align}
\mathbf{u}(k) = \sigma(\mathbf{W}_{2}(\delta(\mathbf{W}_{1}(\mathbf{z})+\mathbf{b}_{1}))+\mathbf{b}_{2}), \mathbf{u} \in \mathbb{R}^{N}
\end{align}

where $\mathbf{W}_{1}$ and $\mathbf{W}_{2}$ are the weight matrices related to two FC layers, $\delta(\cdot)$ is the ReLU function, $\sigma(\cdot)$ is the sigmoid function, and $\mathbf{b}_{1}, \mathbf{b}_{2}$ are the bias of the linear classifiers. Each element of the excitation vector $\mathbf{u}$ emphasizes the corresponding peak feature, which is jointly learned from both the time and frequency domain. Hence $\mathbf{u}$ is called the peak-wise attention vector. Applying the obtained peak-wise attention vector $\mathbf{u}$ to the input feature maps $\mathbf{X}_{tpg}$ and $\mathbf{X}_{fpg}$, we have,

\begin{align}
& \mathbf{X}_{tpg}^{(PA)} = \mathbf{X}_{tpg} \odot \mathbf{u} \\
& \mathbf{X}_{fpg}^{(PA)} = \mathbf{X}_{fpg} \odot \mathbf{u}
\end{align}

Where $\odot$ represents element-wise multiplication in the peak axis. Afterward, we concatenate these feature spaces with peak-attention in the feature axis, which is $\mathbf{Z}_{t}$ for the time-domain and $\mathbf{Z}_{f}$ for the frequency-domain. We get a new feature space $\mathbf{X}_{mp} \in \mathbb{R}^{C \times (Z_{t}+Z_{f}) \times N}$. We pass this feature space through a $2$d convolution layer consists of $D$ number of kernels with $7 \times 7$ kernel size (stride $2$, padding $0$). For our proposed framework, $D$ is $64$. The output from this convolutional filter is $\mathbf{X}_{mp}^{'} \in \mathbb{R}^{D \times h \times w}$. Through this convolutional filter, our model learns jointly from the time and frequency domain features. Afterward, we apply global average pooling on the $\mathbf{X}_{mp}^{'}$ and get vector $\mathbf{d}_{jP} \in \mathbb{R}^{D}$. This vector is passed to an FC layer with $D$ neurons and a sigmoid layer, and we get,

\begin{align}
p_{jp} = \sigma(\mathbf{W}_{jc}(\mathbf{d}_{jp})+b_{jc}), p_{jp} \in \mathbb{R}^{1}
\end{align}

Where, $\mathbf{W}_{jc}$ is the weight matrice and $b_{jc}$ is the bias of the FC layer. This is the probability of that particular signal being damaged. For the negative half-cycle of a particular signal, we get feature spaces with peak-attention $\mathbf{X}_{tng}^{(PA)}$ and $\mathbf{X}_{fng}^{(PA)}$ respectively for time and frequency domain in a similar way as we get for the positive cycle. They are concatenated in the feature axis to get $\mathbf{X}_{mn} \in \mathbb{R}^{C \times (Z_{t}+Z_{f}) \times N}$. The same convolutional filter is applied here, and we get $\mathbf{X}_{mn}^{'}$. Passing it through the same global average pooling, FC layer, and sigmoid layer we get,

\begin{align}
p_{jn} = \sigma(\mathbf{W}_{jc}(\mathbf{d}_{jn})+b_{jc}), p_{jn} \in \mathbb{R}^{1}
\end{align}

\subsubsection*{Classification loss}
\hfill \par
The feature spaces from block-3 of the time-domain branch are $\mathbf{X}_{tpg}$ and $\mathbf{X}_{tng}$. Global average pooling is applied to them, and they are passed to an FC layer with neuron $C$ and sigmoid layer. Similarly, The feature spaces from block-3 for frequency-domain branch are $\mathbf{X}_{fpg}$ and $\mathbf{X}_{fng}$. Global average pooling is applied to them, and they are passed to another FC layer with neuron $C$ and sigmoid layer. Let the vectors after global average pooling is $\mathbf{d}_{tp}, \mathbf{d}_{tn}, \mathbf{d}_{fp},\mathbf{d}_{tn}$ respectively for positive and negative half-cycle of time-domain and frequency-domain.

\begin{align}
& p_{tp} = \sigma(\mathbf{W}_{tc}(\mathbf{d}_{tp})+b_{tc}) \\
& p_{tn} = \sigma(\mathbf{W}_{tc}(\mathbf{d}_{tn})+b_{tc}) \\
& p_{fp} = \sigma(\mathbf{W}_{fc}(\mathbf{d}_{fp})+b_{fc}) \\
& p_{fn} = \sigma(\mathbf{W}_{fc}(\mathbf{d}_{fn})+b_{fc})
\end{align}

Here, $W_{tc},W_{fc}$ are the weight matrices and $b_{tc}, b_{fc}$ are the bias of the linear classifiers. The final classification loss is generated by,

\begin{multline}
\mathcal{L}_{cls} = \mathcal{L}_{bce}(p_{jp}) + \mathcal{L}_{bce}(p_{jn}) + \mathcal{L}_{bce}(p_{tp}) + \mathcal{L}_{bce}(p_{tn}) + \\
\mathcal{L}_{bce}(p_{fp}) + \mathcal{L}_{bce}(p_{fn})  
\end{multline}

$\mathcal{L}_{bce}(\cdot)$ is defined as,
\begin{equation}
\mathcal{L}_{bce}= -\bigg[y\log{(p)}+(1-y)\log{(1-p)}\bigg], 
\end{equation}

Where $p$ is the probability of a sample belonging to a damaged power line or indicating the PD activity and $y$ is the ground truth, $y\subseteq\{0,1\}$. We finally get the total loss by,

\begin{equation}
\mathcal{L}_{total} =\mathcal{L}_{bce}+\lambda \mathcal{L}_{c}
\end{equation}

$\lambda$ is a hyperparameter of our framework and is selected empirically. The proposed framework make the final prediction by simple weighted average of all of these probabilites.

\begin{equation}
p_{final} = \frac{p_{jp}+p_{jn}+p_{tp}+p_{tn}+p_{fp}+p_{fn}}{6}
\end{equation}

\section{Training}
\label{Training}
In this section, we discuss the details of our training procedure. All of our experiments are performed in a hardware environment that includes an Intel Core-i7 7700k, @ 4.20 GHz CPU, and Nvidia GeForce GTX 1070 (8 GB Memory) GPU. All of the necessary codes are written in Python, and we used Pytorch deep learning library \cite{NEURIPS2019_9015} to implement the neural networks.

\begin{table*}[!t]
    \centering
    \caption{Performance comparison of the proposed method with state-of-the-art approaches on the VSB ENET dataset. The best results are shown in \textcolor{Red}{red}.}
    \label{offline_comparision}
    \begin{threeparttable}[b]
    \begin{tabular}{lllll} 
    \hline
    Method & MCC & F1 & Precision & Recall \\
    \hline
    STL + LSTM \cite{Dong2019PartialDD} \tnote{1} & 0.3440 & 0.3582 & 0.2300 & 0.8100  \\
    Random forest \cite{7909221} \tnote{2} & 0.7195 & 0.7783 & 0.8073 & 0.7503 \\
    Resnet18 + VggNet11 \cite{9183469} \tnote{2} & 0.7509 & 0.8432 & 0.8247 & 0.8625 \\
    STL + SVM \cite{8861809} \tnote{1} & 0.7790 & 0.7980 & 0.7300 & 0.8800  \\
    Michau et al. \cite{Michau2021InterpretableDO} \tnote{1} & 0.8170 & 0.8256 & 0.7260 & \textcolor{Red}{\textbf{0.9570}} \\
    Ours & \textcolor{Red}{\textbf{0.8455}} & \textcolor{Red}{\textbf{0.9225}} & \textcolor{Red}{\textbf{0.9357}} & 0.9102 \\
    \hline 
    \end{tabular}
    \begin{tablenotes}
    \item[1] Results reported from the implementation of \cite{Michau2021InterpretableDO}.
    \item[2] Metrics recomputed on our data split assuming
    constant sensitivity and specificity of the model.
    \end{tablenotes}
    \end{threeparttable}
\end{table*}

\begin{table*}[!t]
 \caption{Comparison of the effectiveness of integrating the DDAM with cycle consistency loss. The best results are shown in \textcolor{Red}{red}.}
  \centering
  \begin{adjustbox}{width=\linewidth}
  \begin{tabular}{l|lll|lll|llll}
    \hline
     Experiment Name  & \multicolumn{3}{c|}{PD Signals}  & \multicolumn{3}{c|}{Non-PD Signals} & \multicolumn{4}{c} {Overall} \\
     & Precision & Recall & F1 & Precision & Recall & F1 & Precision & Recall & F1 & MCC \\
     \hline
    TD & 0.7600 & 0.8085 & 0.7835 & 0.9830 & 0.9775 & 0.9802 & 0.8715 & 0.8930 & 0.8819 & 0.7642 \\
    TD+CC & 0.7843 & 0.8511 & 0.8163 & 0.9868 & 0.9794 & 0.9831 & 0.8855 & \textcolor{Red}{\textbf{0.9152}} & 0.8997 & 0.8002 \\
    FD & 0.7333 & 0.7021 & 0.7174 & 0.9705 & 0.9775 & 0.9740 & 0.8519 & 0.8398 & 0.8458 & 0.6933 \\
    FD+CC & 0.7708 & 0.7872 & 0.7789 & 0.9713 & 0.9801 & 0.9757 & 0.8710 & 0.8836 &	0.8773 & 0.7593 \\
    Ensemble & 0.7241 & \textcolor{Red}{\textbf{0.8936}} & 0.8000 & \textcolor{Red}{\textbf{0.9904}} & 0.9700 & 0.9801 & 0.8573 & 0.9318 & 0.8900 & 0.7855 \\
    Dual-CyCon Net & \textcolor{Red}{\textbf{0.8864}} & 0.8298 & \textcolor{Red}{\textbf{0.8571}} & 0.9851 & \textcolor{Red}{\textbf{0.9906}} & \textcolor{Red}{\textbf{0.9878}} & \textcolor{Red}{\textbf{0.9357}} & 0.9102 & \textcolor{Red}{\textbf{0.9225}} & \textcolor{Red}{\textbf{0.8455}} \\
    \hline
  \end{tabular}
  \end{adjustbox}
  \label{Ablation_study_1}
\end{table*}

\subsection{Dataset split}
We follow the same dataset split ratio done by Michau et al. \cite{Michau2021InterpretableDO}. We split the dataset into a training dataset of size 6972, of which 6538 are non-damaged, 434 are damaged power-line samples, and a test dataset of size 1740, of which 1599 are non-damaged, 141 are damaged power-line samples. In our pipeline, we aggregate all pulses from three-phase into a single array. This aggregation results in a training dataset of 2324 samples and a test dataset of 533 samples. We perform the detailed analysis, ablation study, and the impact of the different composing blocks of our proposed framework on this test dataset.

\subsection{Performance metric}
We evaluate the performance of our proposed method based on the Matthews Correlation Coefficient (MCC).

\begin{multline}
MCC=\\
\frac{(TP*TN)-(FP*FN)}{\sqrt{(TP+FP)(TP+FN)(TN+FP)(TN+FN)}}
\end{multline}

Where TP is the number of true positives, TN is the number of true negatives, FP is the number of false positives, and FN is the number of false negatives. Along with MCC, we report the F1, precision, and recall scores of our proposed method for PD signals, non-PD signals, and overall. For a fairer comparison, we follow \cite{Michau2021InterpretableDO} and train our model on a stratified five-folds of the training dataset. We use the best weights from each fold and use a simple average for inference on the test dataset. The best weights are saved based on their performance (MCC) on the validation dataset of each fold.

\subsection{Hyper-parameters}
For ground truth $y$, if any of the three phases of a particular measurement is damaged, we use the ground truth $1$ indicating damaged status. If all three phases are healthy, then we use the ground truth $0$ \cite{Michau2021InterpretableDO}. We use Adam optimizer with standard parameters $\beta_1=0.9$, and $\beta_2=0.99$ \cite{Kingma2015AdamAM}. The batch size is set to $64$. We set the initial learning rate $0.0001$ and run the training for $50$ epochs.

\begin{table*}[t]
 \caption{Comparison of the effectiveness of the attention vector. The best results are shown in \textcolor{Red}{red}.}
  \centering
  \begin{adjustbox}{width=\textwidth}
  \begin{tabular}{l|lll|lll|llll}
    \hline
     Experiment Name  & \multicolumn{3}{c|}{PD Signals}  & \multicolumn{3}{c|}{Non-PD Signals} & \multicolumn{4}{c} {Overall}\\
     & Precision & Recall & F1 & Precision & Recall & F1 & Precision & Recall & F1 & MCC \\
    \hline
    No attention & 0.7500 & \textcolor{Red}{\textbf{0.8936}} & 0.8155 & \textcolor{Red}{\textbf{0.9904}} & 0.9737 & 0.9820 & 0.8700 & \textcolor{Red}{\textbf{0.9336}} & 0.8988 & 0.8014 \\
    Channel attention & 0.7500 & 0.8298 & 0.7879 & 0.9848 & 0.9756 & 0.9802 & 0.8674 & 0.9027 & 0.8840 & 0.7693 \\
    Feature attention & \textcolor{Red}{\textbf{0.9048}} & 0.8085 & 0.8539 & 0.9833 & \textcolor{Red}{\textbf{0.9925}} & \textcolor{Red}{\textbf{0.9879}} & \textcolor{Red}{\textbf{0.9440}} & 0.9005 & 0.9209 & 0.8434 \\
    Peak attention & 0.8864 & 0.8298 & \textcolor{Red}{\textbf{0.8571}} & 0.9851 & 0.9906 & 0.9878 & 0.9357 & 0.9102 & \textcolor{Red}{\textbf{0.9225}} & \textcolor{Red}{\textbf{0.8455}} \\
    \hline
  \end{tabular}
  \end{adjustbox}
  \label{Ablation_study_3}
\end{table*}

\begin{figure*}[!t]
	\centering
	\includegraphics[width=\linewidth]{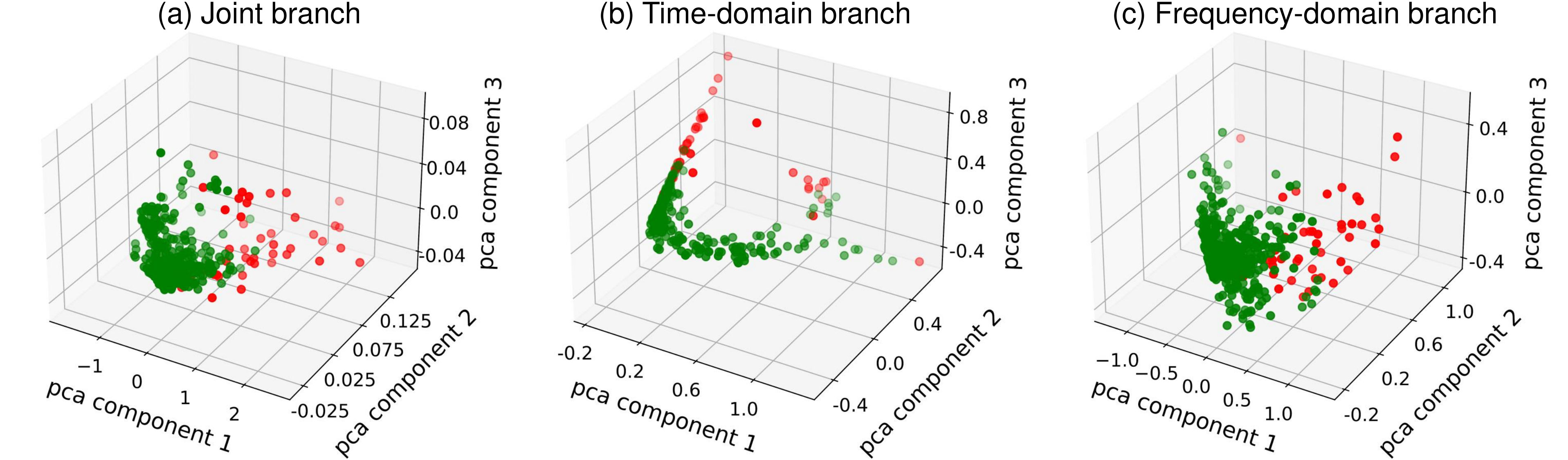}
	\caption{PCA (principal component analysis) visualization of feature distribution of the decision hyper-plane for (a) joint branch (b) time-domain branch (c) frequency-domain branch.}
	\label{pca_plot_1}
\end{figure*}

\section{Experimental Results And Ablation Study}
\label{Experimental Results And Ablation Study}
In this section, we first demonstrate our experimental results on the VSB ENET dataset and compare them with the state-of-the-art results. Afterward, we present the analysis of different components of our proposed framework.

\subsection{Performance on VSB ENET dataset}
We compare our proposed Dual-CyCon Net with previously published state-of-the-art approaches, including the method of \cite{Michau2021InterpretableDO, 8861809, Dong2019PartialDD, 9183469, 7909221}. As we follow the same split ratio as \cite{Michau2021InterpretableDO}, the results of \cite{Michau2021InterpretableDO, 8861809, Dong2019PartialDD} on the test dataset are reported from the implementation of \cite{Michau2021InterpretableDO}. For the results of \cite{9183469, 7909221}, we recompute the value of the metrics they would achieve on our test set, assuming constant sensitivity and specificity of their model. As shown in Table \ref{offline_comparision}, the method of \cite{Michau2021InterpretableDO} is the previous state-of-the-art with an MCC score of 0.8170, while our approach exceeds all the compared methods and achieves a new state-of-the-art performance of 0.8455 MCC. Our classification results outperform others in MCC, F1, and precision categories. In the recall category, our model achieved slightly worse but competitive score. Overall, the proposed Dual-CyCon Net achieves $3.49\%$ and $9.4\%$ relative improvements in MCC and F1 over the former best system by utilizing knowledge from all available domains and exploiting cycle-consistency loss.

\subsection{Effectiveness of Cycle-Consistency loss and DDAM}
The cycle-consistency loss and DDAM are the driving forces of our method that help the model learn the salient knowledge characterizing PD by utilizing time, frequency, and phase information. Both cycle-consistency loss and DDAM work on the high-level feature maps encoded by block-3 of the model backbone. First, we run the experiment with only the time-domain branch, without the frequency-domain branch and DDAM block. We make predictions based only on $p_{tp}$ and $p_{tn}$ and do not use cycle-consistency loss \textbf{(TD)}. Afterward, we run the same setup with cycle-consistency loss \textbf{(TD+CC)}. Similarly, we investigate for the frequency-domain branch without cycle-consistency loss, leaving the time-domain branch and DDAM block \textbf{(FD)}. Then we evaluate the same setup with cycle-consistency loss \textbf{(FD+CC)}. Finally, we run the proposed Dual-CyCon Net. All those results are given in Table \ref{Ablation_study_1} along with the result of the ensemble of the time-domain and frequency-domain models with cycle-consistency losses \textbf{(Ensemble)}. All these experiments are done with cross-validation training, and results are reported on the test dataset. Table \ref{Ablation_study_1} shows that MCC scores improve drastically when the cycle-consistency loss is added, both for time and frequency domain. Also, Dual-CyCon Net achieves a better result than ensemble, proving the efficacy of DDAM block.

\subsection{PCA analysis of the decision hyperplane}
To characterize the decision planes, we take average of $\mathbf{d}_{tp}$, $\mathbf{d}_{tn}$ for time-domain branch, average of $\mathbf{d}_{fp}$, $\mathbf{d}_{fn}$ for frequency domain branch, and average of $\mathbf{d}_{jp}$, $\mathbf{d}_{jn}$ for joint branch. Then, we apply principal component analysis (PCA) of these vectors and resolve them into 3-components. The 3D plots of three PCA components for these three branches are shown in Fig. \ref{pca_plot_1}. In general, as can be seen, the decision planes learned by the model formed clusters of non-PD signals (green) which are separable from the PD (red) clusters.

\subsection{Effectiveness of peak attention vector}
We use Squeeze and Excitation (SE) operation on the peak axis to attain peak attention. To demonstrate the effectiveness, we run the framework without any attention mechanism. We also report the results if the SE operation is done in the channel or feature axis instead of the peak axis. The results are reported in Table \ref{Ablation_study_3}.

\section{Conclusion}
\label{Conclusion}
In this paper, we proposed a new framework for PD activity detection. The proposed method offers several improvements concerning traditional power-line diagnostics. First, our proposed DDAM block can utilize time and frequency domain knowledge cohesively. Our cycle-consistency loss exploits any correlation traces left by PD activity between a particular signal's positive and negative half-cycle. Finally, we provide a detailed analysis of each component of our proposed method. We evaluate our model with other state-of-the-art approaches on a noisy real-world dataset. Our model achieved an MCC score of 0.8455, surpassing the former best performing method.

\balance
\bibliographystyle{IEEEtran}
\bibliography{IEEEabrv,main.bib}

\begin{thebibliography}{10}
\providecommand{\url}[1]{#1}
\csname url@samestyle\endcsname
\providecommand{\newblock}{\relax}
\providecommand{\bibinfo}[2]{#2}
\providecommand{\BIBentrySTDinterwordspacing}{\spaceskip=0pt\relax}
\providecommand{\BIBentryALTinterwordstretchfactor}{4}
\providecommand{\BIBentryALTinterwordspacing}{\spaceskip=\fontdimen2\font plus
\BIBentryALTinterwordstretchfactor\fontdimen3\font minus
  \fontdimen4\font\relax}
\providecommand{\BIBforeignlanguage}[2]{{%
\expandafter\ifx\csname l@#1\endcsname\relax
\typeout{** WARNING: IEEEtran.bst: No hyphenation pattern has been}%
\typeout{** loaded for the language `#1'. Using the pattern for}%
\typeout{** the default language instead.}%
\else
\language=\csname l@#1\endcsname
\fi
#2}}
\providecommand{\BIBdecl}{\relax}
\BIBdecl

\bibitem{7962173}
Y.~Q. Chen, O.~Fink, and G.~Sansavini, ``Combined fault location and
  classification for power transmission lines fault diagnosis with integrated
  feature extraction,'' \emph{IEEE Trans. Ind. Electron.}, vol.~65, no.~1, pp.
  561--569, 2018.

\bibitem{6737537}
H.~K. Agarwal, K.~Mukherjee, and P.~Barna, ``Partially and fully insulated
  conductor systems for low and medium voltage over head distribution lines,''
  in \emph{2013 IEEE 1st International Conference on Condition Assessment
  Techniques in Electrical Systems (CATCON)}, 2013, pp. 100--104.

\bibitem{Dabbak2015SurfaceDC}
S.~Dabbak, H.~Illias, B.~Ang, and M.~Tunio, ``Surface discharge characteristics
  on hdpe, ldpe and pp,'' \emph{Appl. Mech. Mater.}, vol. 785, pp. 383--387, 08
  2015.

\bibitem{Pakonen2007DetectionOI}
\BIBentryALTinterwordspacing
P.~Pakonen, ``Detection of incipient tree faults on high voltage covered
  conductor lines,'' 2007. [Online]. Available:
  \url{http://urn.fi/URN:NBN:fi:tty-200810021011}
\BIBentrySTDinterwordspacing

\bibitem{7962192}
M.~Krátký, S.~Mišák, P.~Gajdoš, P.~Lukáš, R.~Bača, and P.~Chovanec, ``A
  novel method for detection of covered conductor faults in medium voltage
  overhead line systems,'' \emph{IEEE Trans. Ind. Electron.}, vol.~65, no.~1,
  pp. 543--552, 2018.

\bibitem{1038663}
R.~Bartnikas, ``Partial discharges. their mechanism, detection and
  measurement,'' \emph{IEEE Trans Dielectr Electr Insul}, vol.~9, no.~5, pp.
  763--808, 2002.

\bibitem{7422591}
H.~Illias, M.~Tunio, A.~Bakar, H.~Mokhlis, and G.~Chen, ``Partial discharge
  phenomena within an artificial void in cable insulation geometry:
  experimental validation and simulation,'' \emph{IEEE Trans Dielectr Electr
  Insul}, vol.~23, no.~1, pp. 451--459, 2016.

\bibitem{7422593}
F.~Alvarez, J.~Ortego, F.~Garnacho, and M.~Sanchez-Uran, ``A clustering
  technique for partial discharge and noise sources identification in power
  cables by means of waveform parameters,'' \emph{IEEE Trans Dielectr Electr
  Insul}, vol.~23, no.~1, pp. 469--481, 2016.

\bibitem{7480674}
M.~Abd~Rahman, P.~Lewin, and P.~Rapisarda, ``Autonomous localization of partial
  discharge sources within large transformer windings,'' \emph{IEEE Trans
  Dielectr Electr Insul}, vol.~23, no.~2, pp. 1088--1098, 2016.

\bibitem{1678355}
M.~Mashikian and A.~Szarkowski, ``Medium voltage cable defects revealed by
  off-line partial discharge testing at power frequency,'' \emph{IEEE Electr.
  Insul. Mag.}, vol.~22, no.~4, pp. 24--32, 2006.

\bibitem{6957620}
S.~Mišák and V.~Pokorný, ``Testing of a covered conductor’s fault
  detectors,'' \emph{IEEE Trans. Power Deliv.}, vol.~30, no.~3, pp. 1096--1103,
  2015.

\bibitem{1430395}
N.~Sahoo, M.~Salama, and R.~Bartnikas, ``Trends in partial discharge pattern
  classification: a survey,'' \emph{IEEE Trans Dielectr Electr Insul}, vol.~12,
  no.~2, pp. 248--264, 2005.

\bibitem{Hashmi2010ModelingAE}
G.~M. Hashmi, M.~Lehtonen, and M.~Nordman, ``Modeling and experimental
  verification of on-line pd detection in mv covered-conductor overhead
  networks,'' \emph{IEEE Trans Dielectr Electr Insul}, vol.~17, no.~1, pp.
  167--180, 2010.

\bibitem{7909221}
S.~Misák, J.~Fulnecek, T.~Vantuch, T.~Buriánek, and T.~Jezowicz, ``A complex
  classification approach of partial discharges from covered conductors in real
  environment,'' \emph{IEEE Trans Dielectr Electr Insul}, vol.~24, no.~2, pp.
  1097--1104, 2017.

\bibitem{4073965}
Y.~Kawaguchi and S.~Yanabu, ``Partial-discharge measurement on high-voltage
  power transformers,'' \emph{IEEE Trans. Power Appar. Syst.}, vol. PAS-88,
  no.~8, pp. 1187--1194, 1969.

\bibitem{10635_15365}
\BIBentryALTinterwordspacing
J.~JUN, ``Noise reduction and source recognition of partial discharge signals
  in gas-insulated substation,'' 2006. [Online]. Available:
  \url{http://scholarbank.nus.edu.sg/handle/10635/15365}
\BIBentrySTDinterwordspacing

\bibitem{8861809}
M.~Dong, Z.~Sun, and C.~Wang, ``A pattern recognition method for partial
  discharge detection on insulated overhead conductors,'' in \emph{2019 IEEE
  Canadian Conference of Electrical and Computer Engineering (CCECE)}, 2019,
  pp. 1--4.

\bibitem{article_shang}
H.~Shang, J.~Yuan, Y.~Wang, and L.~Zhang, ``Partial discharge pattern
  recognition in power transformer based on multi-kernel multi-class relevance
  vector machine,'' \emph{Diangong Jishu Xuebao/Trans. China Electrotech.
  Soc.}, vol.~29, pp. 221--228, 11 2014.

\bibitem{7757816}
G.~Li, M.~Rong, X.~Wang, X.~Li, and Y.~Li, ``Partial discharge patterns
  recognition with deep convolutional neural networks,'' in \emph{2016
  Condition Monitoring and Diagnosis (CMD)}, 2016, pp. 324--327.

\bibitem{8642226}
B.~Adam and S.~Tenbohlen, ``Classification of multiple pd sources by signal
  features and lstm networks,'' in \emph{2018 IEEE Int. Conf. High Volt. Eng.
  Appl.}, 2018, pp. 1--4.

\bibitem{article_Nguyen}
M.~T. Nguyen, V.-H. Nguyen, S.-J. Yun, and Y.-H. Kim, ``Recurrent neural
  network for partial discharge diagnosis in gas-insulated switchgear,''
  \emph{Energies}, vol.~11, p. 1202, 05 2018.

\bibitem{doi:10.1063/5.0011998}
L.~Wang, K.~Hou, and L.~Tan, ``Research of gis partial discharge type
  evaluation based on convolutional neural network,'' \emph{AIP Adv.}, vol.~10,
  no.~8, p. 085305, 2020.

\bibitem{9087854}
N.~Qu, Z.~Li, J.~Zuo, and J.~Chen, ``Fault detection on insulated overhead
  conductors based on dwt-lstm and partial discharge,'' \emph{IEEE Access},
  vol.~8, pp. 87\,060--87\,070, 2020.

\bibitem{9183469}
W.~Wang and N.~Yu, ``Partial discharge detection with convolutional neural
  networks,'' in \emph{2020 International Conference on Probabilistic Methods
  Applied to Power Systems (PMAPS)}, 2020, pp. 1--6.

\bibitem{Michau2021InterpretableDO}
G.~Michau, C.-C. Hsu, and O.~Fink, ``Interpretable detection of partial
  discharge in power lines with deep learning,'' \emph{Sensors}, vol.~21, p.
  2154, 03 2021.

\bibitem{1430399}
C.~Hudon and M.~Belec, ``Partial discharge signal interpretation for generator
  diagnostics,'' \emph{IEEE Trans Dielectr Electr Insul}, vol.~12, no.~2, pp.
  297--319, 2005.

\bibitem{4446759}
S.~M. Strachan, S.~Rudd, S.~D. McArthur, M.~D. Judd, S.~Meijer, and E.~Gulski,
  ``Knowledge-based diagnosis of partial discharges in power transformers,''
  \emph{IEEE Trans Dielectr Electr Insul}, vol.~15, no.~1, pp. 259--268, 2008.

\bibitem{en14092567}
\BIBentryALTinterwordspacing
O.~Kozák and J.~Pihera, ``Partial discharge analysis and simulation using the
  consecutive pulses correlation method,'' \emph{Energies}, vol.~14, no.~9,
  2021. [Online]. Available: \url{https://www.mdpi.com/1996-1073/14/9/2567}
\BIBentrySTDinterwordspacing

\bibitem{9215344}
K.~M. Mahesh~Kumar, B.~Ramachandra, and L.~S. Kumar, ``Analysis of phase
  resolved partial discharge patterns of kraft paper insulation impregnated in
  transformer mineral oil,'' in \emph{2020 International Conference on Smart
  Electronics and Communication (ICOSEC)}, 2020, pp. 1157--1161.

\bibitem{Altenburger2002AnalysisOP}
R.~Altenburger, C.~Heitz, and J.~Timmer, ``Analysis of phase-resolved partial
  discharge patterns of voids based on a stochastic process approach,''
  \emph{J. Phys. D Appl. Phys.}, vol.~35, p. 1149, 05 2002.

\bibitem{vsb_dataset_manual}
\BIBentryALTinterwordspacing
ENET-Centre, ``Vsb power line fault detection,'' 2019. [Online]. Available:
  \url{https://www.kaggle.com/c/vsb-power-line-fault-detection/}
\BIBentrySTDinterwordspacing

\bibitem{8535878}
Q.~Zhang, J.~Lin, H.~Song, and G.~Sheng, ``Fault identification based on pd
  ultrasonic signal using rnn, dnn and cnn,'' in \emph{2018 Condition
  Monitoring and Diagnosis (CMD)}, 2018, pp. 1--6.

\bibitem{8535913}
K.~Banno, Y.~Nakamura, Y.~Fujii, and T.~Takano, ``Partial discharge source
  classification for switchgears with transient earth voltage sensor using
  convolutional neural network,'' in \emph{2018 Condition Monitoring and
  Diagnosis (CMD)}, 2018, pp. 1--5.

\bibitem{8341662}
H.~Song, J.~Dai, G.~Sheng, and X.~Jiang, ``Gis partial discharge pattern
  recognition via deep convolutional neural network under complex data
  source,'' \emph{IEEE Trans Dielectr Electr Insul}, vol.~25, no.~2, pp.
  678--685, 2018.

\bibitem{HAN2019474}
\BIBentryALTinterwordspacing
T.~Han, C.~Liu, W.~Yang, and D.~Jiang, ``A novel adversarial learning framework
  in deep convolutional neural network for intelligent diagnosis of mechanical
  faults,'' \emph{Knowl Based Syst}, vol. 165, pp. 474--487, 2019. [Online].
  Available:
  \url{https://www.sciencedirect.com/science/article/pii/S0950705118306142}
\BIBentrySTDinterwordspacing

\bibitem{8712431}
H.~Karami, H.~Tabarsa, G.~B. Gharehpetian, Y.~Norouzi, and M.~A. Hejazi,
  ``Feasibility study on simultaneous detection of partial discharge and axial
  displacement of hv transformer winding using electromagnetic waves,''
  \emph{IEEE Trans Industr Inform}, vol.~16, no.~1, pp. 67--76, 2020.

\bibitem{article_Khan}
M.~A. Khan, J.~Choo, and Y.-H. Kim, ``End-to-end partial discharge detection in
  power cables via time-domain convolutional neural networks,'' \emph{J.
  Electr. Eng. Technol.}, vol.~14, 02 2019.

\bibitem{Dong2019PartialDD}
M.~Dong and J.~Sun, ``Partial discharge detection on aerial covered conductors
  using time-series decomposition and long short-term memory network,''
  \emph{Electr. Power Syst. Res.}, vol. 184, p. 106318, 07 2020.

\bibitem{vsb_kaggle_preprocess}
\BIBentryALTinterwordspacing
Mark, 2019. [Online]. Available:
  \url{https://www.kaggle.com/mark4h/vsb-1st-place-solution/}
\BIBentrySTDinterwordspacing

\bibitem{principles_maf}
J.~M. Blackledget, \emph{Digital Signal Processing}.\hskip 1em plus 0.5em minus
  0.4em\relax Woodhead Publishing, 2006.

\bibitem{pmlr-v37-ioffe15}
S.~Ioffe and C.~Szegedy, ``Batch normalization: Accelerating deep network
  training by reducing internal covariate shift,'' in \emph{Proceedings of the
  32nd International Conference on International Conference on Machine Learning
  - Volume 37}, ser. ICML'15.\hskip 1em plus 0.5em minus 0.4em\relax JMLR.org,
  2015, p. 448–456.

\bibitem{8578843}
J.~Hu, L.~Shen, and G.~Sun, ``Squeeze-and-excitation networks,'' in \emph{2018
  IEEE Conference on Computer Vision and Pattern Recognition}, 2018, pp.
  7132--7141.

\bibitem{NEURIPS2019_9015}
A.~Paszke, S.~Gross, F.~Massa, A.~Lerer, J.~Bradbury, G.~Chanan, T.~Killeen,
  Z.~Lin, N.~Gimelshein, L.~Antiga, A.~Desmaison, A.~Kopf, E.~Yang, Z.~DeVito,
  M.~Raison, A.~Tejani, S.~Chilamkurthy, B.~Steiner, L.~Fang, J.~Bai, and
  S.~Chintala, ``Pytorch: An imperative style, high-performance deep learning
  library,'' in \emph{Adv. Neural Inf. Process. Syst. 32}, 2019, pp.
  8024--8035.

\bibitem{Kingma2015AdamAM}
\BIBentryALTinterwordspacing
D.~P. Kingma and J.~Ba, ``Adam: {A} method for stochastic optimization,'' in
  \emph{3rd International Conference on Learning Representations, {ICLR} 2015,
  San Diego, CA, USA, May 7-9, 2015, Conference Track Proceedings}, 2015.
  [Online]. Available: \url{http://arxiv.org/abs/1412.6980}
\BIBentrySTDinterwordspacing

\end{thebibliography}



%








\end{document}